%% file: camera-ready.tex
\documentclass[runningheads]{llncs}
\usepackage{graphicx}

\usepackage{subfigure}
\usepackage{multirow}
\usepackage{booktabs}
\usepackage{bbding}
\usepackage{url}
\usepackage{tikz}
\usepackage{comment}
\usepackage{amsmath,amssymb} % define this before the line numbering.
\usepackage{color}
\usepackage{orcidlink}

\usepackage[accsupp]{axessibility}  % Improves PDF readability for those with disabilities.

\begin{document}
% \linenumbers
\pagestyle{headings}
\mainmatter
\def\ECCVSubNumber{5463}  % Insert your submission number here

\title{E-NeRV: Expedite Neural Video Representation with Disentangled Spatial-Temporal Context} % Replace with your title

% \begin{comment}
% \titlerunning{ECCV-22 submission ID \ECCVSubNumber} 
% \authorrunning{ECCV-22 submission ID \ECCVSubNumber} 
% \author{Anonymous ECCV submission}
% \institute{Paper ID \ECCVSubNumber}
% \end{comment}
%******************

% \begin{comment}
\titlerunning{E-NeRV}
\author{Zizhang Li$^{*}$\orcidlink{0000-0001-9126-7103} \and Mengmeng Wang$^{*}$ \orcidlink{0000-0003-4035-0630}\and Huaijin Pi\orcidlink{0000-0001-7076-1321} \and Kechun Xu\orcidlink{0000-0002-3632-917X} \and Jianbiao Mei\orcidlink{0000-0003-3849-2736} \and Yong Liu$^{\dag}$\orcidlink{0000-0003-4822-8939}}
\authorrunning{Z. Li et al.}
\institute{Zhejiang University \\
\email{\{zzli,mengmengwang,hjpi,kcxu,jianbiaomei\}@zju.edu.cn}\\
\email{yongliu@iipc.zju.edu.cn}}
% \end{comment}
%******************
\maketitle
\let\thefootnote\relax\footnotetext{\scriptsize{$^*$: Equal contributions. $\dag$: Corresponding author.}}

\input{contents/shortcut}

\input{contents/1-abstract}
\input{contents/2-Introduction}
\input{contents/3-related}
\input{contents/4-background}
\input{contents/5-Method}

\input{contents/6-Experiments}
\input{contents/7-conclusion}

% \clearpage\mbox{}Page \thepage\ of the manuscript.
% \clearpage\mbox{}Page \thepage\ of the manuscript.

\clearpage
\bibliographystyle{splncs04}
\bibliography{egbib}
\clearpage

\appendix
\input{contents/8-supp}
\end{document}

%% file: contents/shortcut.tex
\newcommand{\netname}{E-NeRV }

\newcommand{\ba}{\mathbf{a}}\newcommand{\bA}{\mathbf{A}}
\newcommand{\bb}{\mathbf{b}}\newcommand{\bB}{\mathbf{B}}
\newcommand{\bc}{\mathbf{c}}\newcommand{\bC}{\mathbf{C}}
\newcommand{\bd}{\mathbf{d}}\newcommand{\bD}{\mathbf{D}}
\newcommand{\be}{\mathbf{e}}\newcommand{\bE}{\mathbf{E}}
\newcommand{\bff}{\mathbf{f}}\newcommand{\bF}{\mathbf{F}} %
\newcommand{\bg}{\mathbf{g}}\newcommand{\bG}{\mathbf{G}}
\newcommand{\bh}{\mathbf{h}}\newcommand{\bH}{\mathbf{H}}
\newcommand{\bi}{\mathbf{i}}\newcommand{\bI}{\mathbf{I}}
\newcommand{\bj}{\mathbf{j}}\newcommand{\bJ}{\mathbf{J}}
\newcommand{\bk}{\mathbf{k}}\newcommand{\bK}{\mathbf{K}}
\newcommand{\bl}{\mathbf{l}}\newcommand{\bL}{\mathbf{L}}
\newcommand{\bm}{\mathbf{m}}\newcommand{\bM}{\mathbf{M}}
\newcommand{\bn}{\mathbf{n}}\newcommand{\bN}{\mathbf{N}}
\newcommand{\bo}{\mathbf{o}}\newcommand{\bO}{\mathbf{O}}
\newcommand{\bp}{\mathbf{p}}\newcommand{\bP}{\mathbf{P}}
\newcommand{\bq}{\mathbf{q}}\newcommand{\bQ}{\mathbf{Q}}
\newcommand{\br}{\mathbf{r}}\newcommand{\bR}{\mathbf{R}}
\newcommand{\bs}{\mathbf{s}}\newcommand{\bS}{\mathbf{S}}
\newcommand{\bt}{\mathbf{t}}\newcommand{\bT}{\mathbf{T}}
\newcommand{\bu}{\mathbf{u}}\newcommand{\bU}{\mathbf{U}}
\newcommand{\bv}{\mathbf{v}}\newcommand{\bV}{\mathbf{V}}
\newcommand{\bw}{\mathbf{w}}\newcommand{\bW}{\mathbf{W}}
\newcommand{\bx}{\mathbf{x}}\newcommand{\bX}{\mathbf{X}}
\newcommand{\by}{\mathbf{y}}\newcommand{\bY}{\mathbf{Y}}
\newcommand{\bz}{\mathbf{z}}\newcommand{\bZ}{\mathbf{Z}}

\newcommand{\balpha}{\boldsymbol{\alpha}}\newcommand{\bAlpha}{\boldsymbol{\Alpha}}
\newcommand{\bbeta}{\boldsymbol{\beta}}\newcommand{\bBeta}{\boldsymbol{\Beta}}
\newcommand{\bgamma}{\boldsymbol{\gamma}}\newcommand{\bGamma}{\boldsymbol{\Gamma}}
\newcommand{\bdelta}{\boldsymbol{\delta}}\newcommand{\bDelta}{\boldsymbol{\Delta}}
\newcommand{\bepsilon}{\boldsymbol{\epsilon}}\newcommand{\bEpsilon}{\boldsymbol{\Epsilon}}
\newcommand{\bzeta}{\boldsymbol{\zeta}}\newcommand{\bZeta}{\boldsymbol{\Zeta}}
\newcommand{\beeta}{\boldsymbol{\eta}}\newcommand{\bEta}{\boldsymbol{\Eta}} %
\newcommand{\btheta}{\boldsymbol{\theta}}\newcommand{\bTheta}{\boldsymbol{\Theta}}
\newcommand{\biota}{\boldsymbol{\iota}}\newcommand{\bIota}{\boldsymbol{\Iota}}
\newcommand{\bkappa}{\boldsymbol{\kappa}}\newcommand{\bKappa}{\boldsymbol{\Kappa}}
\newcommand{\blambda}{\boldsymbol{\lambda}}\newcommand{\bLambda}{\boldsymbol{\Lambda}}
\newcommand{\bmu}{\boldsymbol{\mu}}\newcommand{\bMu}{\boldsymbol{\Mu}}
\newcommand{\bnu}{\boldsymbol{\nu}}\newcommand{\bNu}{\boldsymbol{\Nu}}
\newcommand{\bxi}{\boldsymbol{\xi}}\newcommand{\bXi}{\boldsymbol{\Xi}}
\newcommand{\bomikron}{\boldsymbol{\omikron}}\newcommand{\bOmikron}{\boldsymbol{\Omikron}}
\newcommand{\bpi}{\boldsymbol{\pi}}\newcommand{\bPi}{\boldsymbol{\Pi}}
\newcommand{\brho}{\boldsymbol{\rho}}\newcommand{\bRho}{\boldsymbol{\Rho}}
\newcommand{\bsigma}{\boldsymbol{\sigma}}\newcommand{\bSigma}{\boldsymbol{\Sigma}}
\newcommand{\btau}{\boldsymbol{\tau}}\newcommand{\bTau}{\boldsymbol{\Tau}}
\newcommand{\bypsilon}{\boldsymbol{\ypsilon}}\newcommand{\bYpsilon}{\boldsymbol{\Ypsilon}}
\newcommand{\bphi}{\boldsymbol{\phi}}\newcommand{\bPhi}{\boldsymbol{\Phi}}
\newcommand{\bchi}{\boldsymbol{\chi}}\newcommand{\bChi}{\boldsymbol{\Chi}}
\newcommand{\bpsi}{\boldsymbol{\psi}}\newcommand{\bPsi}{\boldsymbol{\Psi}}
\newcommand{\bomega}{\boldsymbol{\omega}}\newcommand{\bOmega}{\boldsymbol{\Omega}}

\newcommand{\nA}{\mathbb{A}}
\newcommand{\nB}{\mathbb{B}}
\newcommand{\nC}{\mathbb{C}}
\newcommand{\nD}{\mathbb{D}}
\newcommand{\nE}{\mathbb{E}}
\newcommand{\nF}{\mathbb{F}}
\newcommand{\nG}{\mathbb{G}}
\newcommand{\nH}{\mathbb{H}}
\newcommand{\nI}{\mathbb{I}}
\newcommand{\nJ}{\mathbb{J}}
\newcommand{\nK}{\mathbb{K}}
\newcommand{\nL}{\mathbb{L}}
\newcommand{\nM}{\mathbb{M}}
\newcommand{\nN}{\mathbb{N}}
\newcommand{\nO}{\mathbb{O}}
\newcommand{\nP}{\mathbb{P}}
\newcommand{\nQ}{\mathbb{Q}}
\newcommand{\nR}{\mathbb{R}}
\newcommand{\nS}{\mathbb{S}}
\newcommand{\nT}{\mathbb{T}}
\newcommand{\nU}{\mathbb{U}}
\newcommand{\nV}{\mathbb{V}}
\newcommand{\nW}{\mathbb{W}}
\newcommand{\nX}{\mathbb{X}}
\newcommand{\nY}{\mathbb{Y}}
\newcommand{\nZ}{\mathbb{Z}}

\newcommand{\cA}{\mathcal{A}}
\newcommand{\cB}{\mathcal{B}}
\newcommand{\cC}{\mathcal{C}}
\newcommand{\cD}{\mathcal{D}}
\newcommand{\cE}{\mathcal{E}}
\newcommand{\cF}{\mathcal{F}}
\newcommand{\cG}{\mathcal{G}}
\newcommand{\cH}{\mathcal{H}}
\newcommand{\cI}{\mathcal{I}}
\newcommand{\cJ}{\mathcal{J}}
\newcommand{\cK}{\mathcal{K}}
\newcommand{\cL}{\mathcal{L}}
\newcommand{\cM}{\mathcal{M}}
\newcommand{\cN}{\mathcal{N}}
\newcommand{\cO}{\mathcal{O}}
\newcommand{\cP}{\mathcal{P}}
\newcommand{\cQ}{\mathcal{Q}}
\newcommand{\cR}{\mathcal{R}}
\newcommand{\cS}{\mathcal{S}}
\newcommand{\cT}{\mathcal{T}}
\newcommand{\cU}{\mathcal{U}}
\newcommand{\cV}{\mathcal{V}}
\newcommand{\cW}{\mathcal{W}}
\newcommand{\cX}{\mathcal{X}}
\newcommand{\cY}{\mathcal{Y}}
\newcommand{\cZ}{\mathcal{Z}}

\newcommand{\figref}[1]{Fig.~\ref{#1}}
\newcommand{\secref}[1]{Section~\ref{#1}}
\newcommand{\algref}[1]{Algorithm~\ref{#1}}
\newcommand{\eqnref}[1]{Eq.~\eqref{#1}}
\newcommand{\tabref}[1]{Table~\ref{#1}}

\def\mc{\mathcal}
\def\mb{\mathbf}

\newcommand{\T}{^{\raisemath{-1pt}{\mathsf{T}}}}

\newcommand{\Perp}{\perp\!\!\! \perp}

\makeatletter
\DeclareRobustCommand\onedot{\futurelet\@let@token\@onedot}
\def\@onedot{\ifx\@let@token.\else.\null\fi\xspace}
\def\eg{e.g\onedot} \def\Eg{E.g\onedot}
\def\ie{i.e\onedot} \def\Ie{I.e\onedot}
\def\cf{cf\onedot} \def\Cf{Cf\onedot}
\def\etc{etc\onedot}
\def\vs{vs\onedot}
\def\wrt{wrt\onedot}
\def\dof{d.o.f\onedot}
\def\etal{et~al\onedot}
\def\iid{i.i.d\onedot}
\def\evs{\emph{vs}\onedot}
\makeatother

\newcommand*\rot{\rotatebox{90}}

\newcommand{\boldparagraph}[1]{\vspace{0.2cm}\noindent{\bf #1:} }

\definecolor{darkgreen}{rgb}{0,0.7,0}
\definecolor{lightred}{rgb}{1.,0.5,0.5}

\newcommand{\red}[1]{\noindent{\color{red}{#1}}}
\newcommand{\lightred}[1]{\noindent{\color{lightred}{#1}}}
\newcommand{\ag}[1]{ \noindent {\color{blue} {\bf Andreas:} {#1}} }
\newcommand{\lars}[1]{ \noindent {\color{cyan} {\bf Lars:} {#1}} }
\newcommand{\michael}[1]{ \noindent {\color{blue} {\bf Michael:} {#1}} }
\newcommand{\songyou}[1]{ \noindent {\color{red} {\bf Songyou:} {#1}} }

%% file: contents/1-abstract.tex
\begin{abstract}
Recently, the image-wise implicit neural representation of videos, NeRV, has gained popularity for its promising results and swift speed compared to regular pixel-wise implicit representations. However, the redundant parameters within the network structure can cause a large model size when scaling up for desirable performance. The key reason of this phenomenon is the coupled formulation of NeRV, which outputs the spatial and temporal information of video frames directly from the frame index input. In this paper, we propose E-NeRV, which dramatically expedites NeRV by decomposing the image-wise implicit neural representation into separate spatial and temporal context. Under the guidance of this new formulation, our model greatly reduces the redundant model parameters, while retaining the representation ability. We experimentally find that our method can improve the performance to a large extent with fewer parameters, resulting in a more than $8\times$ faster speed on convergence. Code is available
at \url{https://github.com/kyleleey/E-NeRV}.

\keywords{implicit representation, neural video representation, spatial-temporal disentanglement}
\end{abstract}

%% file: contents/2-Introduction.tex
\section{Introduction}
\label{intro}
Implicit neural representation~(INR) have become popular in recent days. It presents a new manner to represent continuous signals as $f_\theta: \nR^m \rightarrow \nR^n$, which encodes the signal property as a function that maps the $m$-dimensional input~(e.g.\ coordinates) to desired $n$-dimensional output~(e.g.\ RGB values, occupancy, density), and the function is parameterized by deep neural networks with weight $\theta$. Unlike regular grid-wise representations, the compact INRs are proved to be suitable for complex scenes~\cite{mildenhall2020nerf} and arbitrary scale sampling~\cite{chen2021learning}, as well as in lots of 3D tasks~\cite{mildenhall2020nerf,liu2020neural,peng2021neural,oechsle2021unisurf} and image representations~\cite{sitzmann2020implicit,chen2021learning,mehta2021modulated,xu2021ultrasr,xie2021neural,ramasinghe2022regularizing,zhuang2022filtering}. Despite the prevalence of INRs, few works have studied the compatible INR for video signals. 

Video has been treated as an additional supplement of the image in past INR works~\cite{sitzmann2020implicit,mehta2021modulated}. They usually take the $3$-dimensional spatial-temporal coordinate $(x,y,t)$ as input and output RGB values. Most of the following works~\cite{rho2022neural,zhang2021implicit} focusing on video INRs adopt this configuration. However, the training and inference speed of this type of video INR will increase by the order of the third power when processing the video sequences with large resolution and numerous frames. In contrast, a recently proposed method, NeRV~\cite{chen2021nerv}, reformulates the INR of video signals as $f_\theta: \nR\rightarrow\nR^{3\times H\times W}$. Based on the concept that video is a tile of images, NeRV presents an image-wise video INR different from other pixel-wise video INRs. With the frame index in the time axis as input, NeRV directly outputs the desired frame image. The training and inference speed is proved to be much faster than previous methods~\cite{sitzmann2020implicit,tancik2020fourier} experimentally in \cite{chen2021nerv}. And NeRV combines the success of convolution architecture and GAN's network design for its NeRV Blocks, which endows the ability to reconstruct frames of large resolution with high fidelity. By changing the channel dimensions in NeRV Blocks, we can obtain a series of NeRV models with different sizes~(NeRV models with more parameters will naturally perform better). However, as the channel dimensions increase, the model size will increase rapidly. This drawback mainly comes from the architecture of NeRV model, which brings lots of unnecessary and redundant parameters~($2\times$ larger model size when channel dimensions increase 25\%). We ascribe it to the design motivation of NeRV: NeRV considers the spatial and temporal information that lies in each frame image in a hybrid manner and is directly generated from one particular temporal frame index, which results in the heavy model and sub-optimal performance.

\begin{figure}[t]
    \centering
    \includegraphics[width=\linewidth]{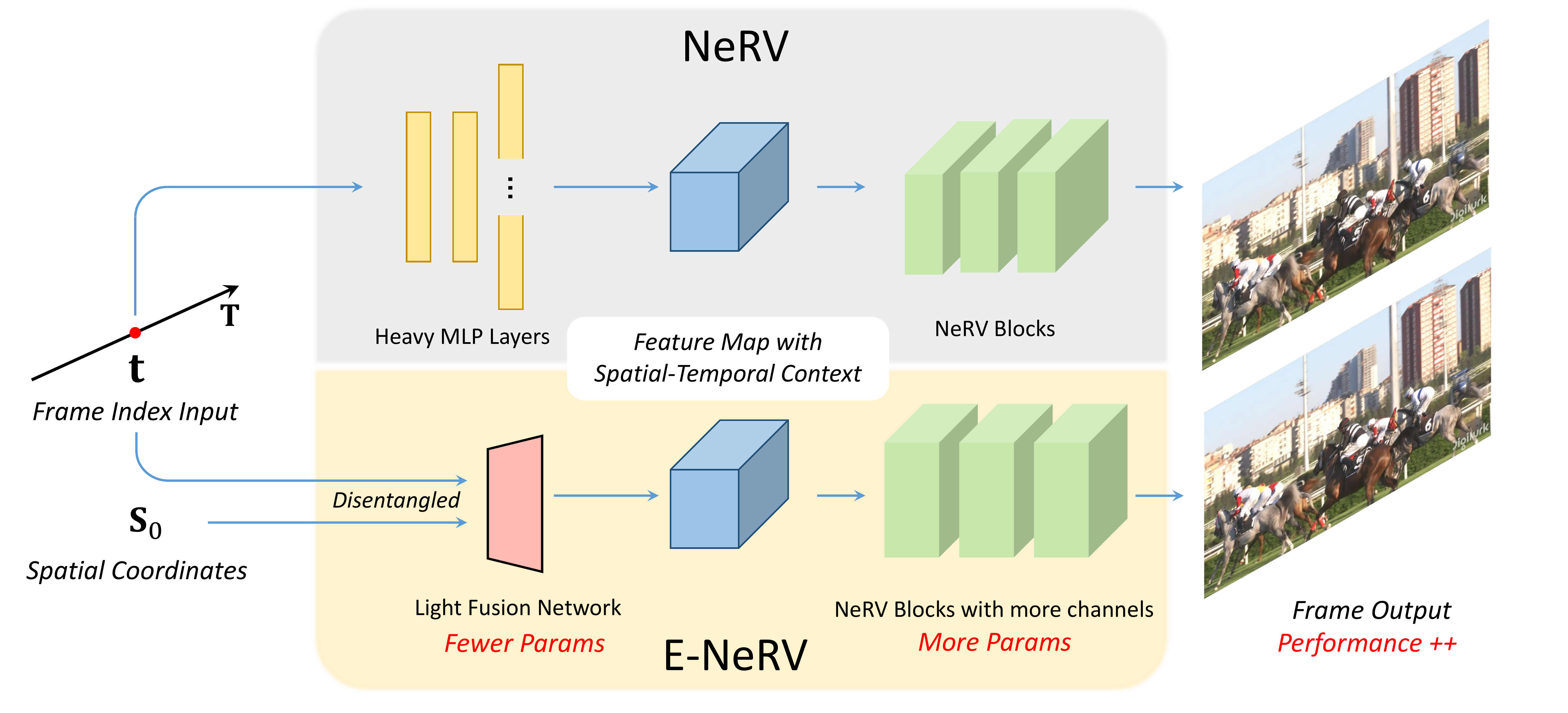}
    \caption{Main motivation of our proposed method. We can greatly reduce the size of the parameters by introducing disentangled spatial-temporal representations with a light network, while maintaining the majority of performance. Furthermore, we distribute the saved parameters to increase channel dimensions in convolution blocks, resulting in an \netname model with similar or fewer parameters but much better performance.}
    \label{fig:motivation}
\end{figure}

Inspired by the video GAN researches~\cite{villegas2017decomposing,hsieh2018learning,yu2021generating} that decompose the content and motion information, we propose the image-wise video INR that explicitly disentangles the spatial-temporal context and fuses them for final prediction, and refactor the original NeRV's network accordingly. Based on this motivation~(illustrated in Fig.~\ref{fig:motivation}), we can effectively lower the parameter size of the baseline model~(from 12M to 5M) while maintaining the majority of performance. We further introduce the temporal embeddings in convolutional blocks to facilitate the representation ability. Besides, we spot the redundant design lying in the NeRV Block and upgrade it. 
We name our method \netname since it \textit{\textbf{E}xpedites} the original NeRV from a disentangled perspective for video implicit representation. We systematically investigate multiple design choices and compare our method with the baseline NeRV model. Our contributions are summarized as follows:
\begin{itemize}
    \item We identify the redundant structures in the image-wise video INR NeRV, which is its major limitation when scaling up for better performance, and attribute this disadvantage to its hybrid formulation.
    \item We propose E-NeRV, a novel image-wise video INR with disentangled spatial-temporal context. 
    \item We demonstrate our method can consistently outperform the NeRV baseline in convergence speed~($8\times$) and performance with fewer parameters. Moreover, the superior performance is consistent among different video INR's downstream applications.
\end{itemize}

%% file: contents/3-related.tex
\section{Related Work}
\label{related}

\subsection{Implicit Neural Representation}

Recently, Implicit Neural Representation~(INR) has gained much popularity for its strong power in modeling a variety of signals. It parameterizes a specific signal by a function that outputs desired properties of the provided coordinate-like input and employs deep neural networks~(usually Multi-Layer Perceptron, MLP) to approximate the function. Thus the signal is implicitly encoded in network's parameters. For instance, the images~\cite{chen2021learning,mehta2021modulated,sitzmann2020implicit} can be defined as RGB values of each pixel location, and 3D objects or scenes can be represented as occupancy~\cite{mescheder2019occupancy,peng2020convolutional}, signed distance~\cite{park2019deepsdf} or radiance field~\cite{mildenhall2020nerf} of each 3D point. INRs are primarily popular in 3D vision tasks like reconstruction~\cite{littwin2019deep,niemeyer2020differentiable,wang2021neus,park2021nerfies,pumarola2021d,li2021neural,xian2021space} and novel-view synthesis~\cite{mildenhall2020nerf,yu2021pixelnerf,barron2021mip,yu2021plenoctrees,tancik2022block}.

\textbf{Implicit Representations of Videos} have not been thoroughly studied in this trend. Regular video implicit representations often take the spatial and temporal index of a pixel, i.e.\ $(x,y,t)\in \bR^3$, as input and output the RGB values of the certain pixel in the certain frame. This simple definition suits short video clips with small image sizes, like $7\times 224\times 224$ in \cite{sitzmann2020implicit,mehta2021modulated}.\cite{rho2022neural} further estimates optical flow for continuous video representation. But this setting is no longer suitable for videos containing hundreds of frames with image resolution at large scale, which requires a long time to optimize and inference~\cite{chen2021nerv} because of the increasing number of frames and pixels. In addition, the paradigm proposed in \cite{mehta2021modulated} for contextual embeddings also can not support the videos with a large amount of frames. Another line of research of video INRs focuses on generative adversarial networks~(GAN)~\cite{goodfellow2014generative}. Instead of generating videos from latent code directly, DiGAN~\cite{yu2021generating} generates parameters of video INRs from context and motion latent code. StyleGAN-V~\cite{skorokhodov2021stylegan} further utilizes convolutional operators for large-scale image synthesis. However, in this work we focus on fitting INR to the particular video instead of generating diverse contents in GAN-based methods.

Recently proposed NeRV~\cite{chen2021nerv} employs an image-wise implicit representation for videos instead of pixel-wise representations before. By combining the implicit representation with advances in convolution for image synthesis, NeRV achieves promising results with less time in training and inference. Following NeRV, our \netname further improves the architecture via a disentangled formulation for a superior performance and fast convergence.

\subsection{Optimization of INRs}

Despite the success of INRs' expression ability, they naturally cost a long time to optimize for considerable performance. Many methods have been proposed to alleviate this problem and also acquire a better representation ability.

From the perspective of function characteristics, researches can be divided into studying an optimal encoding method and applying network regularization. Given that INRs tend to learn better mapping functions with a higher dimensional network input, many following works focus on a better encoding approach. Radial basis function~(RBF)~\cite{dash2016radial} utilizes the weighted sums of embedded RBF encodings. Positional encoding~(PE) proposed in Fourier Feature Networks~(FFN)~\cite{tancik2020fourier} employs a set of Fourier functions to project inputs into high dimensions, and follow-up works~\cite{lin2021barf,hertz2021sape} adopt a coarse-to-fine strategy on frequency for better convergence. Different from using existing functions, SPE~\cite{wang2021spline} uses learnable spline functions and the latest instant-ngp~\cite{muller2022instant} constructs a trainable hash map for shared embedding space. As for the regularization, many consistency constraints~\cite{jain2021putting,deng2021depth,niemeyer2021regnerf} regarding the 3D property have been studied in view synthesis. \cite{tancik2021learned} regularize the in-domain initialization by a meta-learning approach. The distribution-based~\cite{ramasinghe2022regularizing} and Lipschitz-based~\cite{liu2022learning} regularizations can be applied on MLP regarding the smoothness prior for better convergence and generalization.

From the network architecture perspective, some recent works aim to accelerate the training and/or inference of 3D INRs with delicately designed architecture regarding the 3D sparsity. A common approach is to store the feature of MLP inside a pre-defined voxel space~\cite{liu2022learning}, point cloud~\cite{xu2022point} or octree structure~\cite{yu2021plenoctrees}, thus reducing the numbers of point query in both training and inference. To a greater range, SIREN~\cite{sitzmann2020implicit} replaces the commonly used RELU activations in existing MLPs with sinusoidal activation functions and shows the solid fitting ability to complex signals. ACORN~\cite{martel2021acorn} and CoordX~\cite{liang2022coordx} aim at reducing the number of queries to coordinate-based models with different approaches: ACORN~\cite{martel2021acorn} adopts a hierarchical way to decompose the multi-scale coordinates while CoordX~\cite{liang2022coordx} designs a split MLP architecture to leverage the locality between input coordinate points. The following MINER~\cite{saragadam2022miner} improves ACORN via a cross-scale similarity prior.

Our work expedites NeRV from an architecture perspective as we observe the existing unnecessary and redundant structures. By introducing our disentangled formulation, we demonstrate that the resulting model with much fewer parameters can keep the majority of performance or even exceed the NeRV baseline. When scaling up to the same size with baseline NeRV model, our \netname shows greater performance and faster convergence speed.

%% file: contents/4-background.tex
\section{Preliminaries}
\label{Preliminaries}

NeRV~\cite{chen2021nerv}, as an image-wise representation, represents the video as a mapping function $f_\theta: \nR\rightarrow\nR^{3\times H\times W}$ parameterized by the network weight $\theta$. Given a video with $\bT$ frames $\bV=\{\bv_{\bt}\}_{\bt=1}^{\bT}$, the input is a scalar frame index value which is normalized to $\bt\in\left[0, 1\right]$, and the output is the whole corresponding frame image $\bv_{\bt}\in\nR^{3\times H\times W}$. By taking a closer look at its architecture, the formulation can be split into two parts:

\begin{equation}
\begin{aligned}
    &\bff_{\bt} = \texttt{RESHAPE}\left(F\left(\gamma\left(\bt\right)\right)\right) \in \nR^{C\times h\times w}, \\
    &\bv_{\bt} = G\left(\bff_{\bt}\right).
    \label{formula:NeRV}
\end{aligned}
\end{equation}
The $\gamma\left(\bt\right)$ means the regular frequency positional encoding proposed in \cite{mildenhall2020nerf}:

\begin{equation}
    \gamma\left(\bt\right) = (\sin(b^0\pi\bt), \cos(b^0\pi\bt), \dots, \sin(b^{l-1}\pi\bt), \cos(b^{l-1}\pi\bt)),
    \label{formula:pe}
\end{equation}
where $b$ and $l$ are hyper-parameters. The function $F$ stands for the MLP while the function $G$ stands for the convolutional generator. To be more specific, it contains a sequence of NeRV Blocks with convolution and pixel-shuffle layers for the up-sample and image generation purpose. The network first maps the positional encoding of input frame index to a 1-$\bd$ feature vector and then reshapes the vector to a 2-$\bd$ feature map $\bff_\bt\in \nR^{C\times h\times w}$, where $(h, w)=(9, 16)$ in NeRV's setting. The following convolution and pixel-shuffle operation gradually transform the feature map to the original image size. And the $1\times1$ convolution with sigmoid generates the desired three channels of normalized RGB values.

The success of NeRV comes from several reasons. It employs an image-wise representation, which avoids per-pixel training and inference. The quantitative comparison in \cite{chen2021nerv} shows great training and inference speed improvement compared to pixel-wise representations. The NeRV Block containing convolution and pixel-shuffle is suitable for image generation and leads to about 40 PSNR of final performance, superior to other video implicit neural representations.

A series of models with different sizes and performances are provided in \cite{chen2021nerv}. A larger model can obtain better performance, and the way to scale up the model size is to increase the channel dimensions within the NeRV Blocks.
However, this paradigm remains drawbacks. First is the last layer of MLP. To generate a feature vector that can be reshaped to feature map of size $C\times h\times w$~($\sim 10^5$), the last layer of MLP can be extensive, and some naive solutions will cause a large performance drop~(details in Section.~\ref{alternative exp}). Then, the convolution kernel can also be vast because of the following large-scale factor pixel-shuffle layer. NeRV considers the image-wise video implicit representation as an index-to-image formulation, while we consider it as a generation process with disentangled formulation, and the frame index only represents the temporal context. In Section.~\ref{method}, we elaborate our attempt to upgrade the redundant structure with spatial-temporal disentanglement, and we quantitatively and qualitatively show the significant performance and convergence speed of our method in Section.~\ref{experiments}.

%% file: contents/5-Method.tex
\section{Methodology}
\label{method}

\begin{figure}[t]
    \centering
    \includegraphics[width=\linewidth]{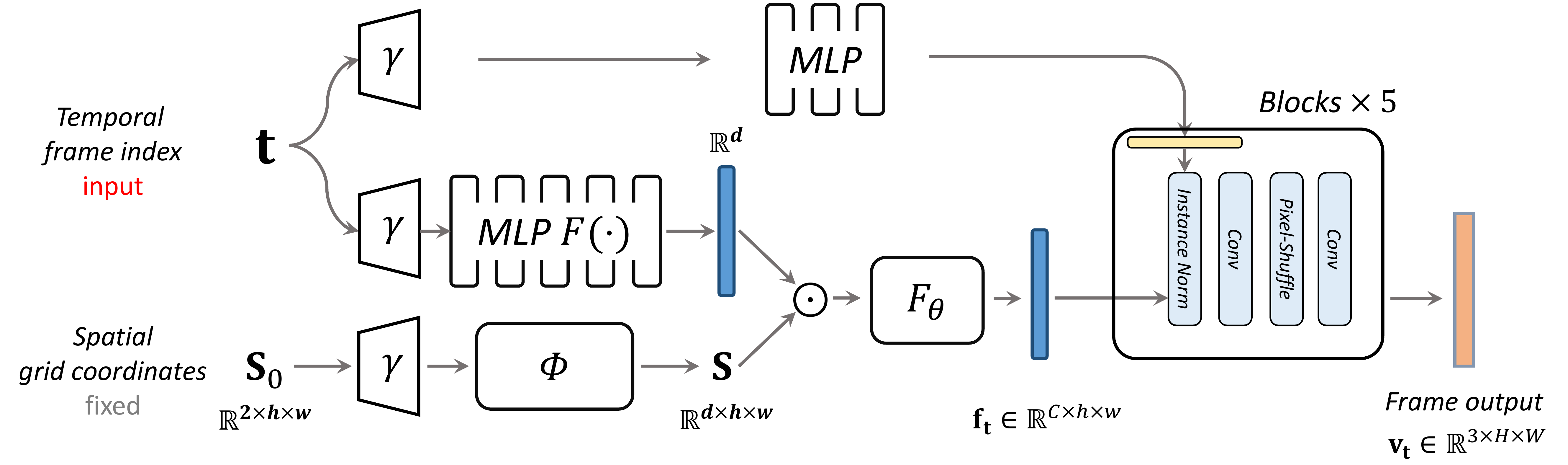}
    \caption{Architecture of proposed E-NeRV. Our spatial-temporal feature map $\bff_\bt$ is generated from disentangled spatial and input temporal contexts with fewer parameters~(Section.~\ref{frame condition}). The temporal information is also introduced to the convolution stages as a normalization procedure~(Section.~\ref{frame condition}) for better performance. In addition, we re-design the NeRV Blocks to further remove the redundant structures~(Section.~\ref{bottleneck}).}
    \label{fig:arch}
\end{figure}

The overall architecture of the proposed \netname is illustrated in Fig.~\ref{fig:arch}. This section will introduce our approach towards the redundant parameters and structures. More specifically, in Section.~\ref{frame condition} we state how to disentangle spatial and temporal representation and the resulting formulation and architecture. And in Section.~\ref{bottleneck} we elaborate on our upgraded design of NeRV block. 
% Further in Section.~\ref{temporal freq}, we analyze how encoding frequency in our method unaffectedly improves the temporal interpolation smoothness.

\subsection{Disentangled image-wise video INR}
\label{frame condition}

The first redundant part in NeRV emerges at the last layer of MLP. For instance, the NeRV-L model with $12.5$M parameters, almost $70\%$ of its size comes from the last MLP layer, which outputs $\bff_\bt\in \nR^{112\times 9\times 16}$. Although the height and width of the feature map are relatively small, it requires large channel numbers to guarantee the final performance. In experiments~(Section.~\ref{alternative exp}), we show some trivial modifications that may ease the large size of parameters, but lead to a dramatic performance drop compared to ours. We claim that this structure needs to exist because NeRV generates frame feature map $\bff_\bt$ directly and only from the input $\bt$, which means to derive the spatial and temporal information \textit{together} from the temporal input. 

As an alternative, we propose to disentangle the spatial-temporal information, and let temporal input become a feature vector to manipulate over the spatial space. In detail, we reformulate the generation of $\bff_\bt$ as follows:

\begin{equation}
    \bff_\bt = F_{\theta}(F(\gamma(\bt))\odot\bS).
    \label{eq:disentagle_formulaton}
\end{equation}

Here $F$ still stands for an MLP network, but with a much less parameter size because the output of $F$ in our method is only a $d$-dimensional vector, where $d\ll C\times h\times w$. We decompose the spatial-temporal information into the temporal one encoded in $F(\gamma(\bt))$, and the spatial one encoded in the spatial context embeddings $\bS\in\nR^{d\times h\times w}$. Then a lightweight network $F_\theta$ is employed to fuse the separated spatial and temporal information into the spatial-temporal embeddings.

Since $\bS$ is expected to contain the spatial context, we initialize it using the normalized grid coordinates. Accordingly we get the initialized spatial context $\bS_0\in \nR^{2\times h\times w}$. First, we encode the $\bS_0$ into $\hat{\bS}_0$ using similar frequency positional encoding $\gamma(\cdot)$ in Eq.~\ref{formula:pe}. Then, we adopt a small transformer~\cite{vaswani2017attention} with single-head self-attention and residual connection here to encourage the feature fusion among spatial locations to get the spatial context $\bS$:

\begin{equation}
\begin{aligned}
    \bS &=\Phi(\hat{\bS}_0) = \texttt{softmax}(\bq^T\bk)\bv + \hat{\bS}_0 \\ 
        &= \texttt{softmax}(f_q(\hat{\bS}_0)^T f_k(\hat{\bS}_0))f_v(\hat{\bS}_0) + \hat{\bS}_0.
\end{aligned}
\label{eq:spatial_feature}
\end{equation}
where $f_{*}$ stands for different projection networks to project input feature map's channel dimension to desired dimension $d_t$. $\bq,\bk,\bv$ denote for the query, key and value of the transformer. Now the $\bS$ can be considered as embeddings containing the desired spatial context. And when representing different videos, the learnable parameters in $\Phi$ are different. In other words, we parameterize the spatial information in a video in the weights of $\Phi$.
%, and $\gamma(\cdot)$ is conducted at first for each $\bS_0$ in Eq.~\ref{eq:spatial_feature}, which we omit for clearness

Next, after the disentangled procedure, we need to fuse the temporal vector $F(\gamma(\bt))\in\nR^d$ with the spatial context $\bS\in\nR^{d\times h\times w}$  to obtain the spatial-temporal information. First, we element-wise multiply the temporal vector with each feature vector from all the locations in $\bS$. Then, we utilize $F_\theta$ to further fuse the features together. $F_\theta$ here can be any operations as long as it can encourage spatial and channel feature fusion. We employ a tiny multi-head attention transformer network similar to $\Phi$ for its ability of long-range modeling and feature fusion. In experiments, we further compare this choice with other alternatives~(Section.~\ref{alternative exp}). 
% It's worth noting that any complicated operations able to fuse features together may serve as the $F_\theta$ here.

Besides, we observe that temporal information in NeRV is only related to the feature map at the beginning of function $G$ in Eq.~\ref{formula:NeRV}. Therefore, we further fuse temporal context to each NeRV Blocks in $G$ to make sufficient and thorough use of the temporal embedding. In experiments we find this design can further boost the performance. In detail, we take inspiration from the design of GAN~\cite{huang2017arbitrary}, and consider the temporal context as a concept of style vector. Unlike using element-wise multiplication to get coarse spatial-temporal feature map, here the temporal information only plays a role of distribution shift. As illustrated in the upper part of Fig.~\ref{fig:arch}, we adopt a tiny MLP~($\sim$ 0.2M) to generate temporal feature $\bl_\bt\in\nR^{d_0}$. Then for the $i$-th block~($i=1,\dots, 5$), a linear layer $M_i$ generates per-channel mean $\mu_i$ and standard deviation $\sigma_i$ accordingly. We denote the input feature map for the $i$-th block as $\bff_\bt^i$. This newly generated distribution shifts the feature map as an instance normalization with temporal context:

\begin{equation}
    IN(\bff_\bt^i) = \sigma_i\left(\frac{\bff_\bt^i - \mu(\bff_\bt^i)}{\sigma(\bff_\bt^i)}\right) + \mu_i
\end{equation}
where $\mu(\bff_\bt^i)$ and $\sigma(\bff_\bt^i)$ are computed across spatial dimensions. This operation is conducted at the beginning of each block to let the temporal information guide the generation of the corresponding frame.

\subsection{Upgraded NeRV Block}
\label{bottleneck}

As stated in Section~\ref{Preliminaries}, another redundant structure lies in NeRV block. Because the convolution needs to generate enough channels for further pixel-shuffle operation, if the input feature map's channel dimension is $C_1$, desired output dimension is $C_2$, the up-sample scale factor is $s$ and kernel size is $3\times 3$, regardless of the bias, the size of trainable weight is $C_1\times C_2\times s\times s\times 3\times 3$. When scale factor $s$ is large, for example, $s=5$ in the first NeRV block, the size can be enormous~(up to $65\%$ of the overall model) if we scale up the input and output channel dimension for better performance.

In order to tackle this problem, we modify the NeRV block with a subtle design: we replace the convolution kernel with two consecutive convolution kernels with small channels. Then we place the pixel-shuffle operation in the middle and introduce an intermediate dimension $C_0$. By using \texttt{conv}$\left(\cdot,\cdot\right)$ to denote convolution kernel with corresponding input and output channel dimensions, our new architecture can be formulated as:

\begin{equation}
\texttt{conv}(C_1, C_0\times s\times s)\rightarrow\texttt{pixel-shuffle}(s)\rightarrow\texttt{conv}(C_0, C_2),
\end{equation}
and the parameters in this new formula are: $3\times 3\times C_0 \times(C_1 \times s\times s + C_2)$.

In practice, we set $C_0 = \texttt{min}(C_1, C_2) / 4$. If $C_1 \leq C_2$, the ratio of the parameters size is $(C_1 / 4C_2 + 1 / 4s^2)\approx C_1 / 4C_2 \leq 1/4$. We find replacing the first NeRV Block with this design can bring largely simplified size while maintaining most of the performance~(see Section.~\ref{exp-process}). The reason is that the scale factor of the first block equals $5$ and thus results in an oversize model. The following blocks with factor equaling $2$ will not benefit much from this modification, so in our final setting, we replace the first NeRV block with our upgraded version.

%% file: contents/6-Experiments.tex
\section{Experiments}
\label{experiments}

\subsection{Datasets and Implementation Details}
\label{datasets and implementation}

We conduct quantitative and qualitative comparison experiments on $8$ different video sequences collected from scikit-video and UVG~\cite{mercat2020uvg} datasets, similar to experiment settings in \cite{chen2021nerv}. Each video sequence contains about $150$ frames and with a resolution of $1280\times 720$.

We set up-scale factors $5, 2, 2, 2, 2$ for each block of our model to reconstruct a $1280\times 720$ image from the feature map of size $16\times 9$. We follow the training schedule of the original NeRV implementation for a fair comparison. We train the model using Adam optimizer~\cite{kingma2014adam}. Each model is trained for $300$ epochs on each video sequence unless specified, with the batchsize of $1$. 

We adopt NeRV-L with 12.57M parameters as our baseline. For the part in our model that is orthogonal to our modification, we follow the same settings as in NeRV, like activation choice. We set $d=d_t=256$ for spatial and temporal feature fusion, $d_0=128$ for temporal instance normalization. We set all the positional encoding layers in our model identical to NeRV's positional encoding formulated in Eq.~\ref{formula:pe}, and we use $b=1.25$ and $l=80$ if not otherwise denoted. For training objective, we use the same combination of L1 and SSIM loss as \cite{chen2021nerv}:

\begin{equation}
    L = \frac{1}{T}\sum_{t=1}^T \alpha ||\bv_t-\hat{\bv}_t||_1 + (1-\alpha)(1-\texttt{SSIM}(\bv_t, \hat{\bv}_t)).
\end{equation}

The $\alpha$ is set to $0.7$, $T$ stands for the total number of frames, $\bv_t$ denotes the reconstructed frame image while $\hat{\bv}_t$ denotes its corresponding ground truth. Please refer to the supplementary material for more implementation details, experiments, results and visualizations.

% \red{datasets, implementation detail regarding the stride and pe parameters}

\subsection{Process of Removing Redundant Part and Scaling Up}
\label{exp-process}

\begin{figure}[t]
    \centering
    \includegraphics[width=\linewidth]{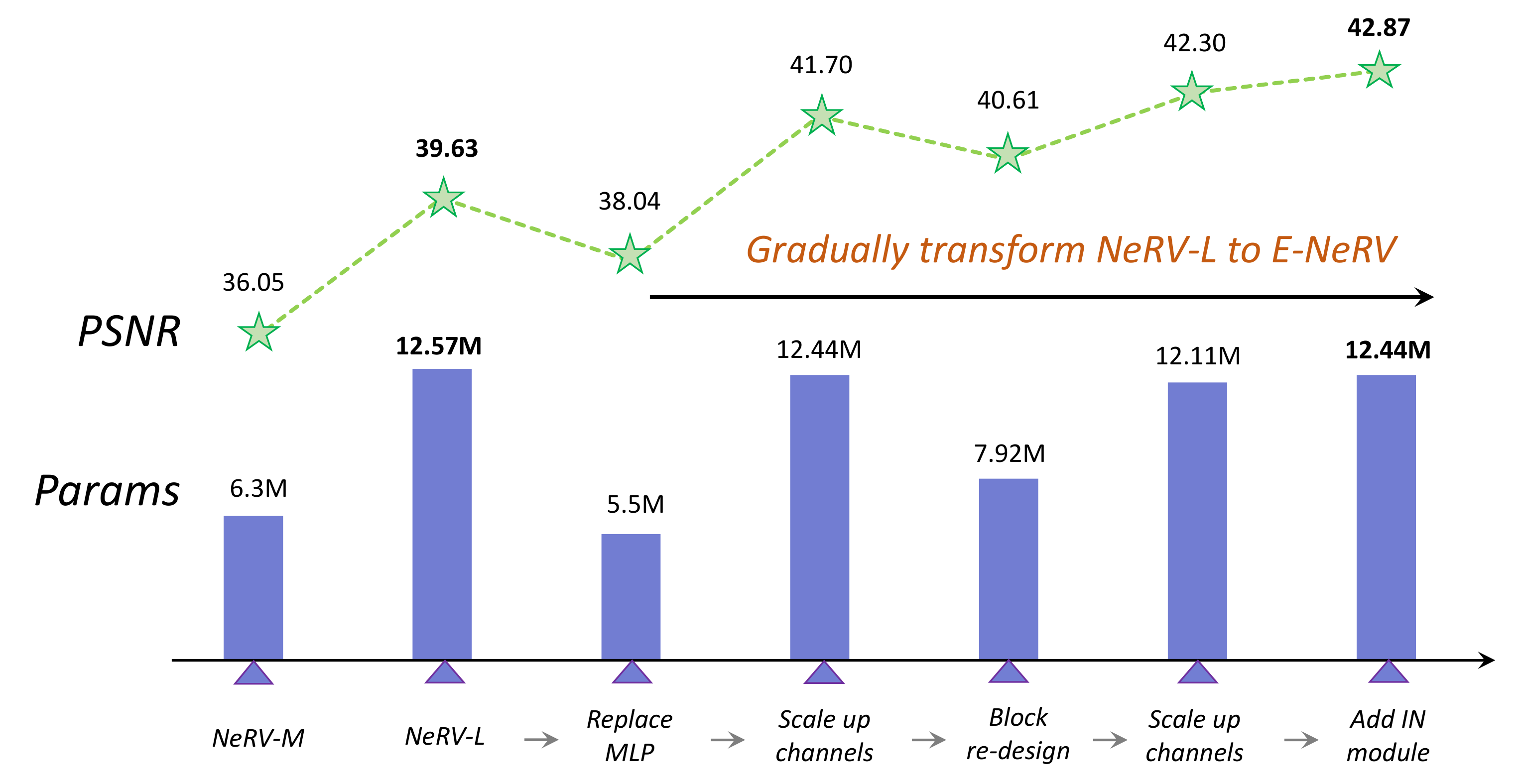}
    \caption{Process of gradually transforming original 12.57M NeRV-L with 39.63 PSNR to our \netname with slightly fewer parameters but much better performance. PSNR results are tested on ``Bunny'' video. Please see Section.~\ref{exp-process} for detailed descriptions.}
    \label{fig:exp-parameters-process}
    
\end{figure}

In this section, we show how to replace the redundant structures and parameters with our proposed methods, and gradually distribute the saved parameters to the channels in convolution stages, which leads to \netname with fewer parameters but much better performance at last.

The overall process is shown in Fig.~\ref{fig:exp-parameters-process}. We first replace the heavy MLP with our disentangled formulation in Eq.~\ref{eq:disentagle_formulaton} and corresponding structure. This step can decrease the parameters from 12.57M to 5.5M, while the obtained model can still get 38.04 PSNR. As a comparison, NeRV-M model in \cite{chen2021nerv} with more parameters can only reach a much worse performance of 36.05 PSNR. Then we first scale up the channels in convolution blocks for a model with a size similar to NeRV-L, and the model after scaling can get 41.70 PSNR.

After the first scaling, another redundant structure emerges: the NeRV Block with up-scale factor $5$ and large channel dimension can be overwhelming, so we replace it with our new design. As shown in Fig.~\ref{fig:exp-parameters-process}, the resulting model reduces $37\%$ parameters. It is notable that the obtained model already has less parameters~(7.92M \emph{vs}.\ 12.57M) but better performance~(40.61M \emph{vs}.\ 39.63M) compared to origination NeRV-L. Then we scale up channels again and add temporal instance normalization branch at last for our proposed E-NeRV.

\subsection{Main Results}
% \red{schedule comparison \& performance comparison given same size, on all videos}
\begin{table*}[t]
    \centering
    \setlength{\tabcolsep}{0.7mm}
     \caption{PSNR~(larger the better) comparison between NeRV-L and our method given similar model size under the same training schedule. The last row indicates performance improvement brought by our method. Our method consistently outperforms the baseline model on diverse kinds of video sequences.}
    \begin{tabular}{llccccccccc}
    \toprule
     {} & {} & & Bunny & Beauty & Bosphorus & Bee & Jockey & SetGo & Shake & Yacht  \\
     \midrule
     NeRV-L& 12.57M & & 39.63 & 36.06 & 37.35 & 41.23 & 38.14 & 31.86 & 37.22 & 32.45 \\
     \midrule
     \multirow{2}{*}{Ours} & \multirow{2}{*}{12.49M} & & 42.87 & 36.72 & 40.06 & 41.74 & 39.35 & 34.68 & 39.32 & 35.58 \\
     \cmidrule{4-11}
     {} & {} & & \textbf{$\uparrow$ 3.24} & \textbf{$\uparrow$ 0.66} & \textbf{$\uparrow$ 2.71} & \textbf{$\uparrow$ 0.51} & \textbf{$\uparrow$ 1.21} & \textbf{$\uparrow$ 2.82} & \textbf{$\uparrow$ 2.10} & \textbf{$\uparrow$ 3.13} \\
     \bottomrule
     
\end{tabular}
    \label{tab:main_result_300ep_psnr}
\end{table*}

We provide a comparison of our method and NeRV in Table.~\ref{tab:main_result_300ep_psnr}. We refer to \cite{chen2021nerv} for further comparison to pixel-wise video INRs such as SIREN~\cite{sitzmann2020implicit} and FFN~\cite{mildenhall2020nerf}, which demonstrates that NeRV surpasses these methods in both performance and speed. Although our proposed \netname has similar speed and parameters, it consistently performs better than the NeRV on different video sequences.

Because the design of our proposed \netname does not employ any kinds of data prior, we claim this improvement exists when using \netname to represent any video sequences. Notably, our method can bring larger promotion for the videos with more dynamic contents in Table~\ref{tab:main_result_300ep_psnr}, for example, the ``Bunny'' and ``Yacht'' videos. We assume this is because our disentangled implicit representation can better model the spatial and temporal variations for the videos with more dynamic contents.

\begin{figure}[t]
\centering
    \subfigure[PSNR on ``Bunny'' video]{
        \begin{minipage}[b]{0.45\linewidth}
        \centering
            \includegraphics[width=\linewidth]{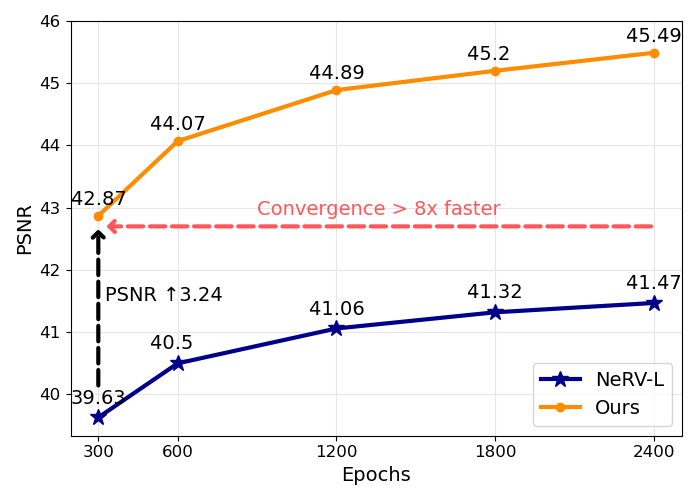}
            
            % \caption{}
            \label{fig:psnr_epochs_bunny}
        \end{minipage}
    }
    \subfigure[PSNR on ``Yacht'' video]{
        \begin{minipage}[b]{0.45\linewidth}
        \centering
            \includegraphics[width=\linewidth]{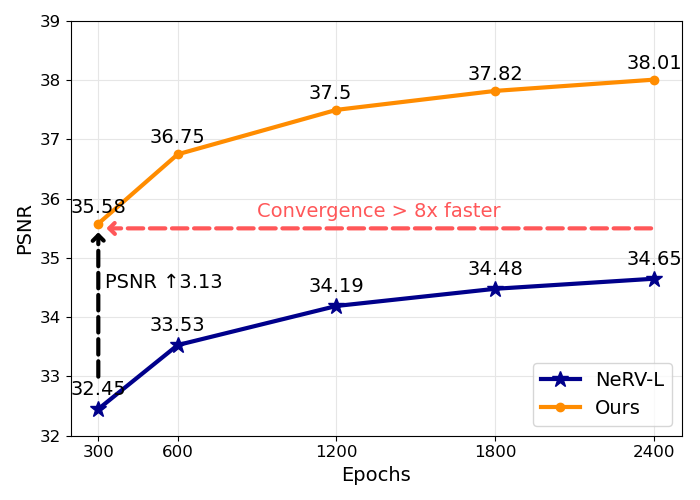}
           
            % \caption{}
            \label{fig:psnr_epochs_yacht}
        \end{minipage}
    }
    \caption{PSNR \emph{vs}.\ Epochs. Comparison of NeRV-L and our method on ``Bunny'' and ``Yacht'' videos. Our method's performance within $300$ epochs is better than NeRV's at $2400$ epochs, which shows better performance and faster convergence.}
    
    \label{fig:psnr_epochs}
\end{figure}

Since training the INRs to fit a video sequence is an over-fitting process, longer schedule naturally leads to better performance. In other words, if the proposed method's performance surpasses another method with the same schedule, it guarantees a better performance and faster convergence speed simultaneously. In Fig.~\ref{fig:psnr_epochs}, we provide a comparison between our method and NeRV on ``Bunny'' and ``Yacht'' videos with different training schedules. Our method's performance at $300$ epochs exceeds the baseline's with a large margin. It also surpasses the baseline's performance at $2400$ epochs, as $8\times$ faster on convergence. Actually, our method's performances at $300$ epochs beat the baseline's at $2400$ epochs on all the diverse videos. We provide the detailed results in supplementary.

\subsection{Comparison with Alternatives}
\label{alternative exp}

We compare our method with four alternative approaches attempting to remove the redundant parameters or conduct fusion of $F_\theta$ in Eq.~\ref{eq:disentagle_formulaton}:

\boldparagraph{NeRV-$C_S$} Since the last layer of MLP with an output size of $C\times h\times w$ causes overwhelming parameters, we add an intermediate channel dimension $C_S$ which is lower than $C$. The MLP outputs feature map at size $C_S\times h\times w$, and a following $1\times1$ convolution will increase the channel dimension to $C$ as original setting before the NeRV Blocks.

\boldparagraph{NeRV-Split} Inspired by split architecture in \cite{liang2022coordx}, we redesign the MLP structure and let it output the tensor with size $C\times(h + w)$, then split it into two parts with size $C\times h$ and $C\times w$ respectively. A tensor product is conducted to generate the desired $C\times h\times w$ feature map $\bff_\bt$ accordingly.

\boldparagraph{E-NeRV-MLP} Since the function $F_\theta$ is responsible for feature fusion of spatial and temporal context, any fusion operation is applicable. We replace our original setting, a small transformer block with attention mechanism, with two successive MLPs on spatial channels~($h\times w$) and feature channels~($C$).

\boldparagraph{E-NeRV-Conv} We use $3\times 3$ convolution block to replace the transformer block. The convolution block fuses the features within a window region and scans over the entire feature map in a sliding window manner.

The results are shown in Table.~\ref{tab:alternative}. For fair comparison on how to lower the size of the parameter, we establish two versions of our method: we remove the part of the structure to introduce the temporal context in convolution blocks stage as described in \ref{frame condition} since it can further boost the performance, and decrease the convolution's channel dimensions to make the resulting models' parameter sizes identical to the sizes of two alternatives. It can be seen that our method surpasses these alternatives given similar parameters settings.

% GATHER TWO TABLES
\begin{table}[t]
\centering
    \begin{minipage}{0.4\linewidth}
        \centering
        \caption{Alternative comparison}
        \resizebox{\textwidth}{!}{%
            \begin{tabular}{lcccc}
            \toprule
             {} & Size & PSNR $\uparrow$ & MS-SSIM $\uparrow$  \\
             \midrule
             NeRV-$C_S$& 5.8M & 33.72 & 0.9562 \\
             \midrule
             Ours-1\dag  & 5.8M & 35.91 & 0.9738 \\
             \midrule
             NeRV-Split& 7.2M & 35.78 & 0.9724 \\
             \midrule
             Ours-2\dag & 7.2M & 36.74 & 0.9782 \\
             \midrule
             E-NeRV-MLP & 12M & 38.54 & 0.9861 \\
             \midrule
             E-NeRV-Conv & 12.5M & 38.67 & 0.9865 \\
             \midrule
             \netname & 12.5M & 38.79 & 0.9866 \\
             \bottomrule
        \end{tabular}
        }
        
        \label{tab:alternative}
        \end{minipage}
    \begin{minipage}{0.48\linewidth}
    \centering
        \caption{Ablation study on components}
        \resizebox{\textwidth}{!}{
            \begin{tabular}{lccccc}
            \toprule
             {} & $\Phi$ & $F_\theta$ & IN & PSNR $\uparrow$ & MS-SSIM $\uparrow$  \\
             \midrule
             NeRV-L & - & - & - & 36.74 & 0.9802 \\
             \midrule
             Variant 1 & \XSolidBrush & \XSolidBrush & \XSolidBrush & 37.89 & 0.9837 \\
             \midrule
             Variant 2 & \Checkmark & \XSolidBrush & \XSolidBrush & 38.09 & 0.9843 \\
             \midrule
             Variant 3 & \Checkmark & \Checkmark & \XSolidBrush & 38.45 & 0.9855 \\
             \midrule
             \netname & \Checkmark & \Checkmark & \Checkmark & 38.79 & 0.9866 \\
             \bottomrule
        \end{tabular}
        }
        \label{tab:ablation}
    \end{minipage}
    
\end{table}

As for the alternatives for feature fusion in $F_\theta$, transformer can bring incremental performance growth compared with MLP or Conv. However, all three models can beat NeRV-L with a large margin. The disentangled representation and structure itself can significantly lower the size, so that we can distribute the saved parameters to the convolution for much better performance. With the rapid growth of vision transformer research~\cite{han2020survey}, any other more complicated structures, like the combination of transformer and convolution, are also welcomed and may further raise the performance. We claim that on some videos with almost still content, like ``Beauty'' and ``Bee'', the differences between each alternative are slight compared to the improvement on more dynamic videos. Since metrics are averaged over all videos, the difference between alternatives may also seem incremental in Table.~\ref{tab:alternative}, but the partial ordering relation is the same over $8$ videos.

\subsection{Ablation Studies}
\label{ablation}

In this section, we study the effects of three novel components of our proposed method: the spatial fusion function $\Phi$ at the beginning of the network, the spatial and temporal fusion $F_\theta$ and the temporal instance normalization method to introduce temporal context in each convolution blocks. The ablation experiments are executed on all the video sequences and obtained metrics are averaged.

As shown in Table.~\ref{tab:ablation}, \netname obtains better performance as gradually adding these modules, and this increasing property exists on all the experiment video sequences. It is notable that the ``Variant 1'', without the fusion and temporal context in convolution stages, can still outperform baseline on both metrics among different video sequences. To be more specific, simply using the proposed disentanglement formulation to reduce the redundant parameters and distributing them to the following convolution blocks, the obtained model with similar parameters can already surpass the NeRV-L. We claim this further proves the effectiveness of our disentanglement motivation to some extent.

\subsection{Downstream application results}
\begin{table*}[t]
	\centering
	\renewcommand\arraystretch{1.4}
\caption{
		Denoising(\textbf{left}) and compression(\textbf{right}) results comparison of our method and NeRV.
	}
	\begin{tabular}{cc}
	\resizebox{0.2\textwidth}{!}{
    \centering
			\input{contents/figs/denoising}} &
		\begin{minipage}{0.65\textwidth}
        \centering
			\includegraphics[width=0.85\textwidth]{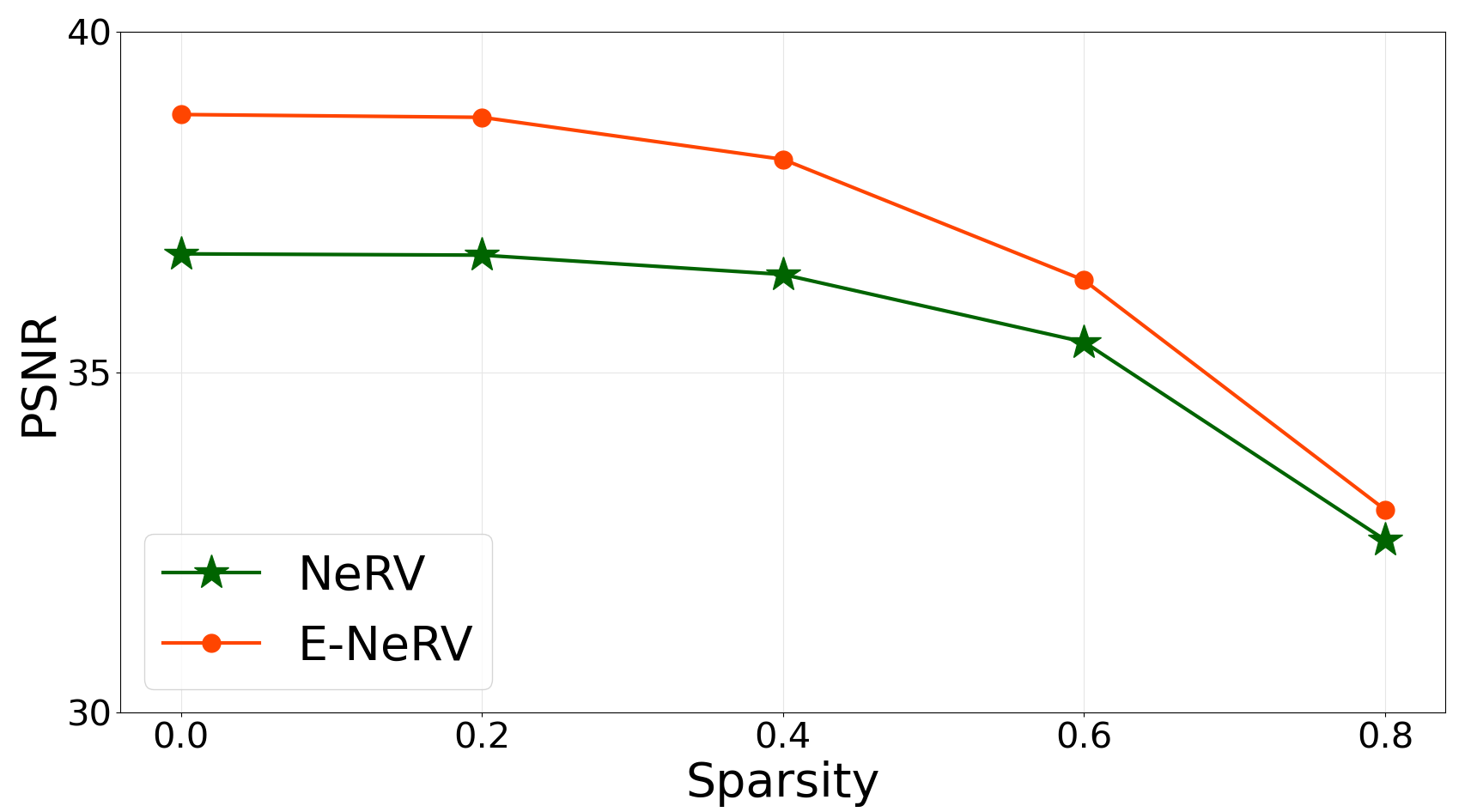}
		\end{minipage}
	\end{tabular}
	\label{tab:downstream}
\end{table*}
In addition to the representation ability, we also compare the E-NeRV's performance to NeRV's on different downstream tasks for video INR, including video denoising and compression. The results are shown in Table.~\ref{tab:downstream}. 

Both experiments follow the NeRV's pipeline and we further conduct an ablation on different prune ratios for compression. The PSNR metrics are average among all the video sequences. In denoising results the ``Noise" refers to noisy frames before any denoising. Here we only compare to NeRV since they beat other filter-based and learning-based methods in their paper. The denoising results of E-NeRV also prove the advantage of our disentangled spatial representation, which serves as a spatial prior in video denoising.

For the compression experiments, the performances of both methods drop as increasing the compression ratio~(Sparsity in the figure), but E-NeRV remains better performance at all different compression ratios. The results also show that the compression ability of frame-wise video INR, i.e.\ the pipeline of pruning the network weight as compressing the video sequence, remains intact for the proposed E-NeRV. The detailed overall results can be found in the supplementary.

\begin{figure}[t]
    \centering
    \includegraphics[width=\linewidth]{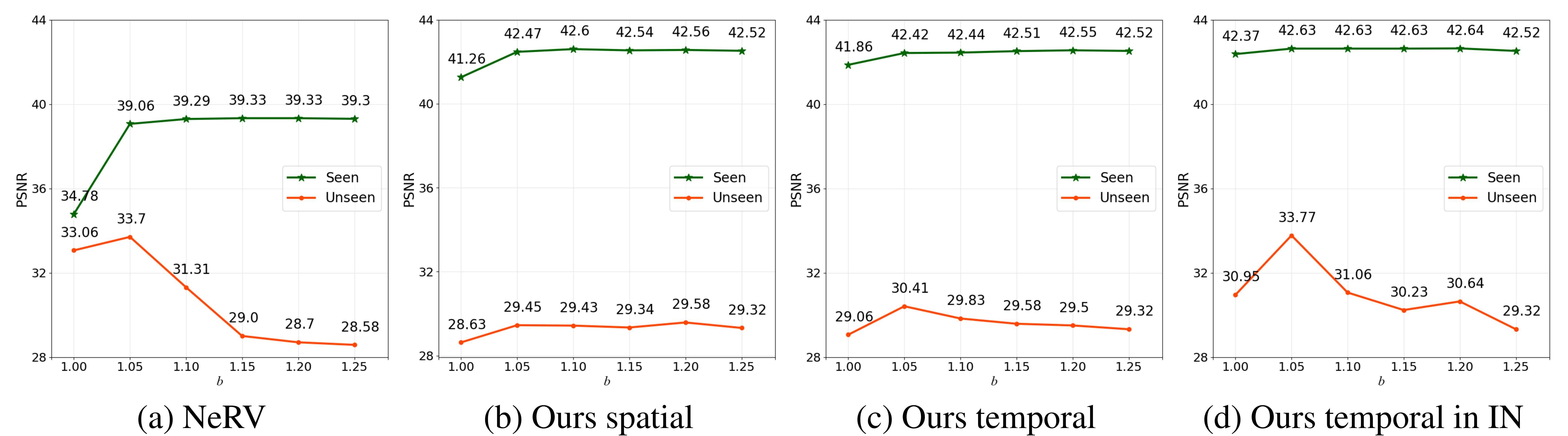}
    \caption{Frequency variation for different encodings of: (a)~NeRV's input $\bt$, (b)~our spatial map $\bS_0$, (c)~input $\bt$ in E-NeRV and (d)~$\bt$ for IN.}
    \label{fig:freq_small_exp}
\end{figure}

\subsection{Temporal frequency analysis}
\label{temporal_freq_exp}

The frequency of Fourier feature mapping can greatly influence the INR's representation ability~\cite{tancik2020fourier}. A minor frequency may lead to smoothness among input and suitable for interpolation but also degrade INR's fitting on training points.

In this section, we study the effect of different frequency in our disentangled representation. We devide the video at a ratio of $3 : 1$ into seen and unseen frames, and adjust the frequency which is $1.25$ in our general setting.  The results are provided in Fig.~\ref{fig:freq_small_exp}.

Starting from NeRV's interpolation~($39.3/28.58$) at frequency $1.25$, we can see that since NeRV consider spatial and temporal in a coupled manner, lowering the frequency can boost the interpolation but also cause a performance drop on seen frames~(Fig.~\ref{fig:freq_small_exp}~(a)). On the contrary, our disentangled representation allows manipulating frequency in three encodings: spatial grid coordinates, temporal input $\mathbf{t}$ and $\mathbf{t}$ used in temporal IN. In detail, adjusting the frequency in IN module from $1.25$ to $1.05$ leads to the optimal interpolation while perserving the performance on training points~(Fig.~\ref{fig:freq_small_exp}~(d)), which can be considered as another advantage from our disentangle structure. More dataset partition details, interpolation results and visualization are available in supplementary.

%% file: contents/figs/denoising.tex
\begin{tabular}{lc}
    \toprule
    {} &  PSNR \\
    \midrule
    Noise  & 28.60\\
    NeRV     & 34.69\\
    E-NeRV&     \textbf{36.23}\\
    \bottomrule
\end{tabular}

%% file: contents/7-conclusion.tex
\section{Conclusion}
\label{conclusion}

In this paper we present E-NeRV, an image-wise video implicit representation with disentangled spatial and temporal context. Following previous image-wise video INR~\cite{chen2021nerv}, our method retains its advantages on training and inference speed compared to pixel-wise video INRs~\cite{tancik2020fourier,sitzmann2020implicit,mehta2021modulated}, but boosts the performance and convergence speed with a large margin. We quantitatively show that our proposed disentanglement structures together with other modifications can greatly reduce the original unnecessary and redundant parameters. By reallocating the saved parameters, our method with fewer parameters can perform much better, with an $8\times$ faster convergence speed. We experimentally analyze the function of each component in our method on diverse video sequences.

Finally, we remark our method can be further improved by applying a more effective and sophisticated feature fusion method for our disentangled representations. In future work, we plan to apply our image-wise video INR to other downstream tasks like optical flow estimation and video super-resolution.
~\\

\noindent \textbf{Acknowledgments.}~We thank all authors and reviewers for the contributions. This work is supported by a Grant from the National Natural Science Foundation of China~(No.~U21A20484).

%% file: contents/8-supp.tex
\section{Additional Implementation Details}

\subsection{Datasets}

\begin{figure}[htbp]
    \centering
    \includegraphics[width=\linewidth]{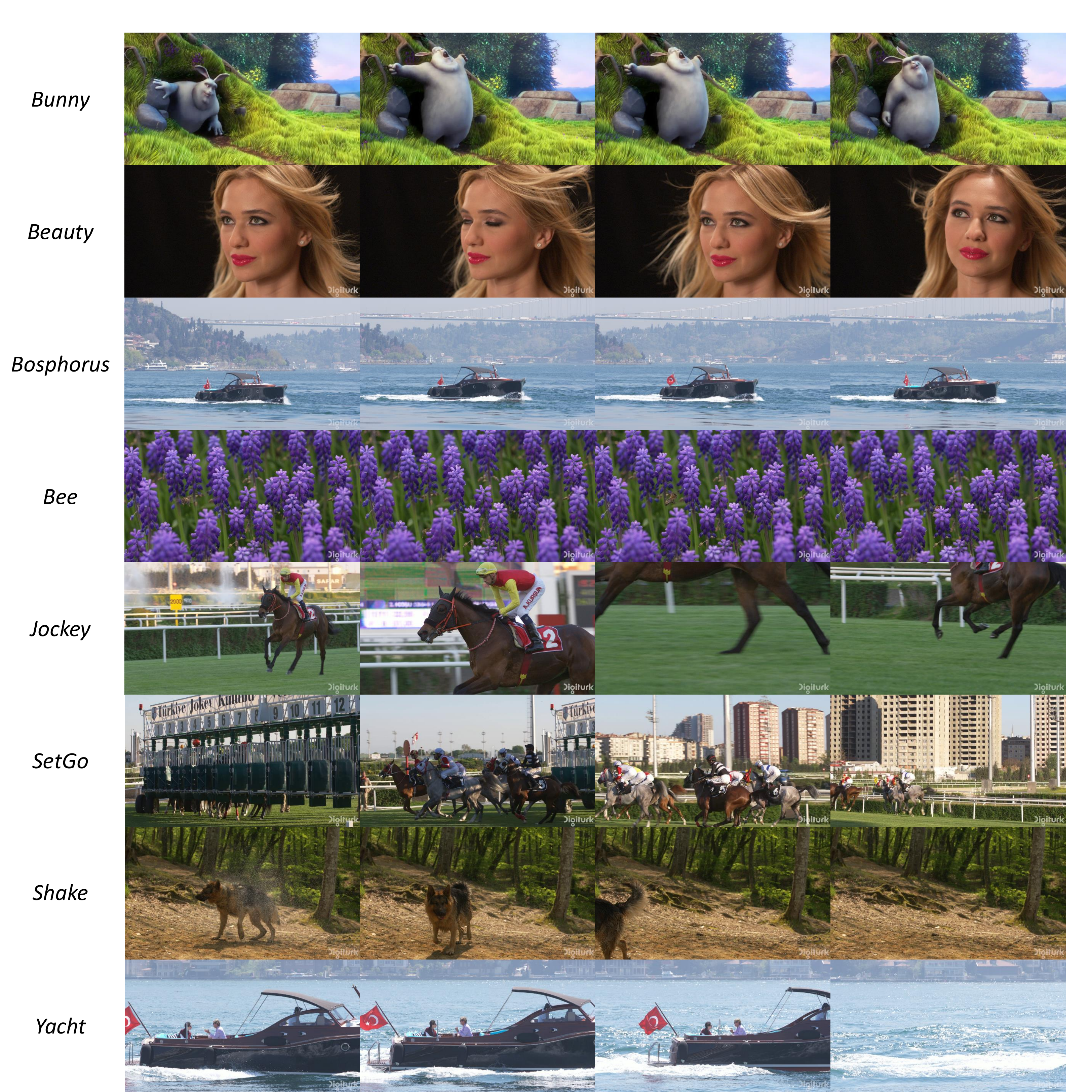}
    % \rule{12cm}{4cm}
    \caption{A visualization of frame images in the video sequences that been used in experiments. Name of each sequence is indicated on the left.}
    \label{fig:dataset}
\end{figure}

We use eight different video sequences to conduct experiments for representation ability. Here we provide a visualization of some random selected frame images from these videos in Fig.~\ref{fig:dataset}. It can be seen that the video sequences we used have diverse types of contents, and our method outperforms baseline method on all of these sequences.

For the dataset split in Section.~\ref{temporal_freq_exp}, we split each sequence at a ratio of $3:1$. To be more specific, for every four consecutive frames, the last frame is distributed into the ``unseen'' split while the other three frames are assigned to ``seen'' split. The normalized frame index is calculated before the partition, so we can use the frame index in ``unseen'' split to quantitatively experiment the temporal interpolation ability.

\subsection{Network Architecture}
\label{supp-arch}

\begin{figure}[h]
    \centering
    \includegraphics[width=\linewidth]{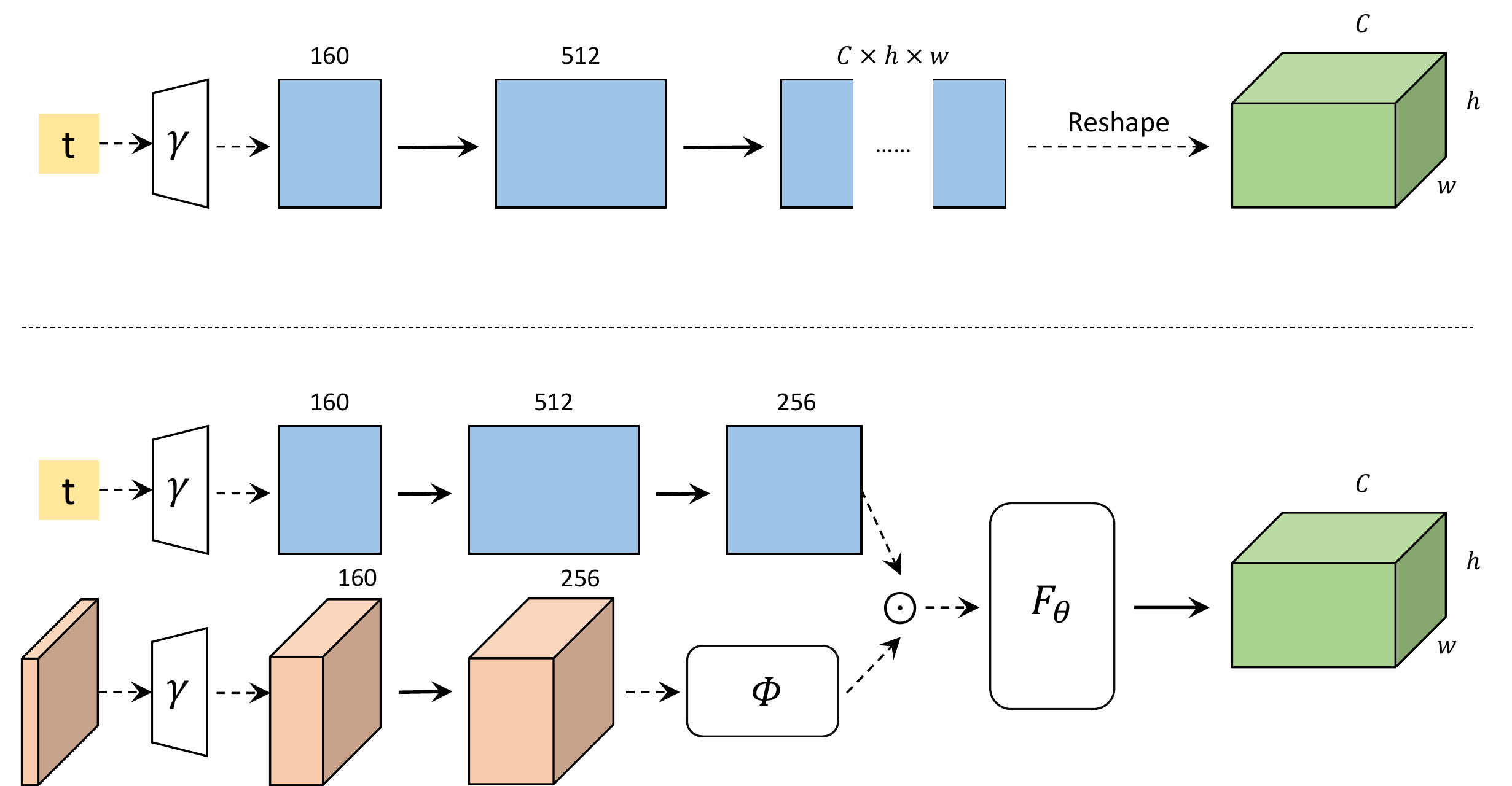}
    \caption{Comparison of how to generate spatial-temporal feature map in NeRV and our method. The black solid arrows stand for MLP layers with learnable parameters. Our method greatly reduce the size of parameters in this stage by introducing the disentangled structure. Detailed description of $\Phi$ and $F_\theta$ is in \ref{supp-arch}.}
    \label{fig:comparison_first}
\end{figure}

\begin{figure}[h]
    \centering
    \includegraphics[width=\linewidth]{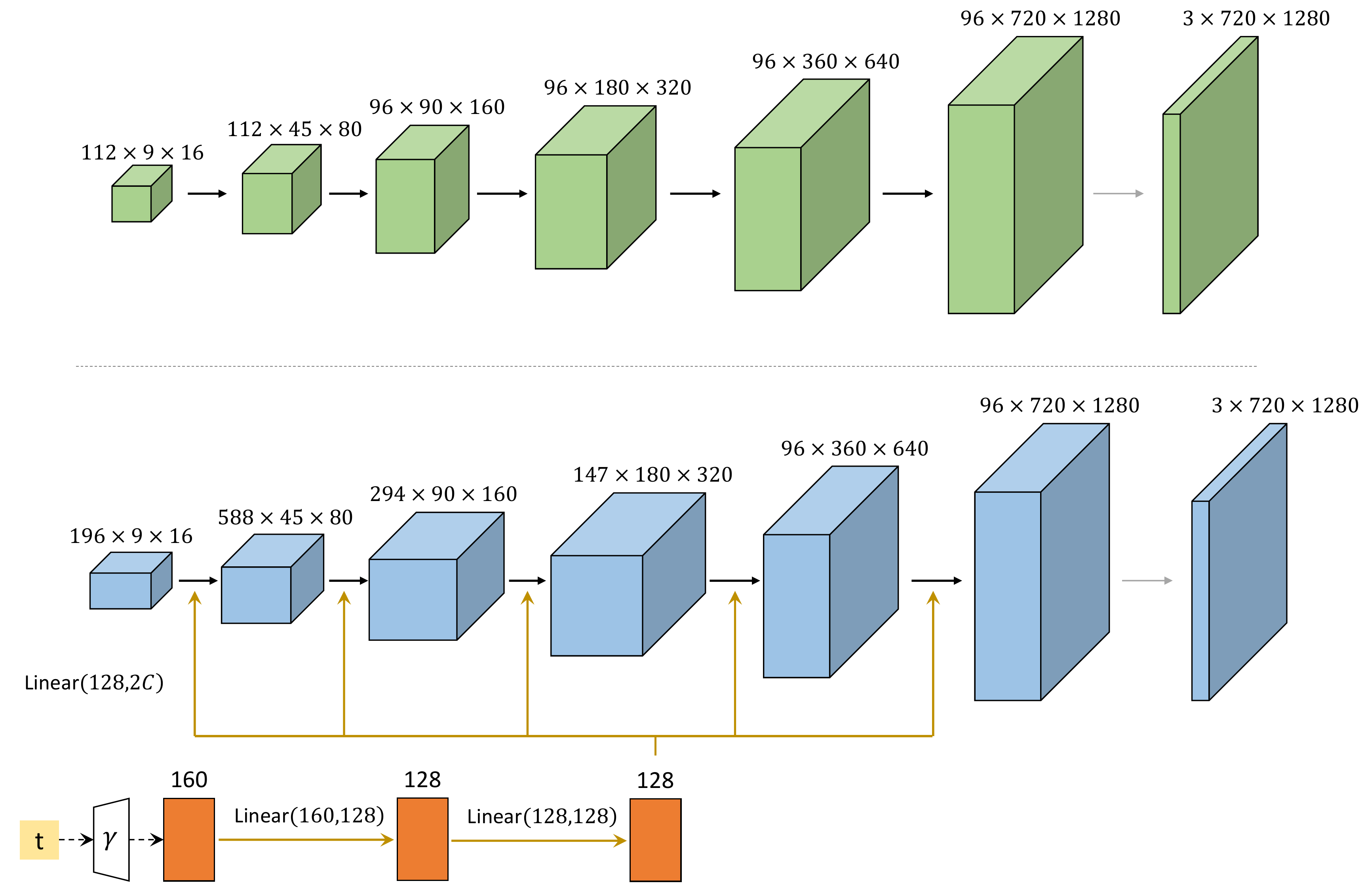}
    \caption{Comparison of the convolution stage, and also our proposed temporal instance normalization module. The black solid arrows stand for $3\times 3$ convolution, pixel-shuffle and activation to up-scale the feature map. We adopt our upgraded design~(Section.~\ref{bottleneck}) at the first block. The gray arrows stand for $1\times 1$ convolution with a sigmoid function to get the desired 3-channel image. The brown arrows indicate the MLP layers for temporal instance normalization. Since we lower the size of first stage, in this stage we can increase the channel dimensions for much better performance.}
    \label{fig:comparison_second}
\end{figure}

We provide a detailed illustration of the architectures of $12.49$M E-NeRV and $12.57$M NeRV-L. Splitting the network structure into two parts: before and after the generation of feature map $\bff_\bt$, in Fig.~\ref{fig:comparison_first} we show the comparison of structure and parameters of original coupled MLP and our proposed disentangled network. The input and output dimensions of single-head-attention transformer block $\Phi$ is 256, and the dimension of MLP layer inside the $\Phi$ is 128. The transformer block $F_\theta$ is almost identical to $\Phi$ except it supports a multi-head-attention where number of heads equals 8. In Fig.~\ref{fig:comparison_second} we show the comparison between the similar convolution stages and also our proposed branch for temporal instance normalization.

\subsection{Training Details}

For fair comparison, we trained all of our models following the schedule proposed in \cite{chen2021nerv}. The maximum of learning rate is set to be 5e-4 and follows the ``warmup-cosine-drop'' schedule. The learning rate increases linearly in first $20\%$ of the iterations and drops to 0 in a cosine schedule in the rest of the iterations. As mentioned in \cite{chen2021nerv} that more iterations lead to better performance, we set the batchsize of 1 in all the experiments to maximize the nubmers of back propagation.  All experiments are run with NVIDIA RTX2080ti, and since the batchsize is small, GPU with less memory can also support. We implement all of our experiments in PyTorch.

\subsection{Metrics}

For the representation ability, we adopt two metrics in our experiments: PSNR and MS-SSIM. These two metrics all measure the similarity between the ground truth frame and the frame image that network outputs. The PSNR stands for ``Peak Signal-to-Noise Ratio'', calculated on the basis of mean-square error of two images. The MS-SSIM stands for ``Multiscale structural similarity'', it is calculated on the basis of luminance, contrast and structure. The method with larger value of both metrics have better performance on reconstructing the frame image.

\begin{figure}[t]
    \centering
    \includegraphics[width=\linewidth]{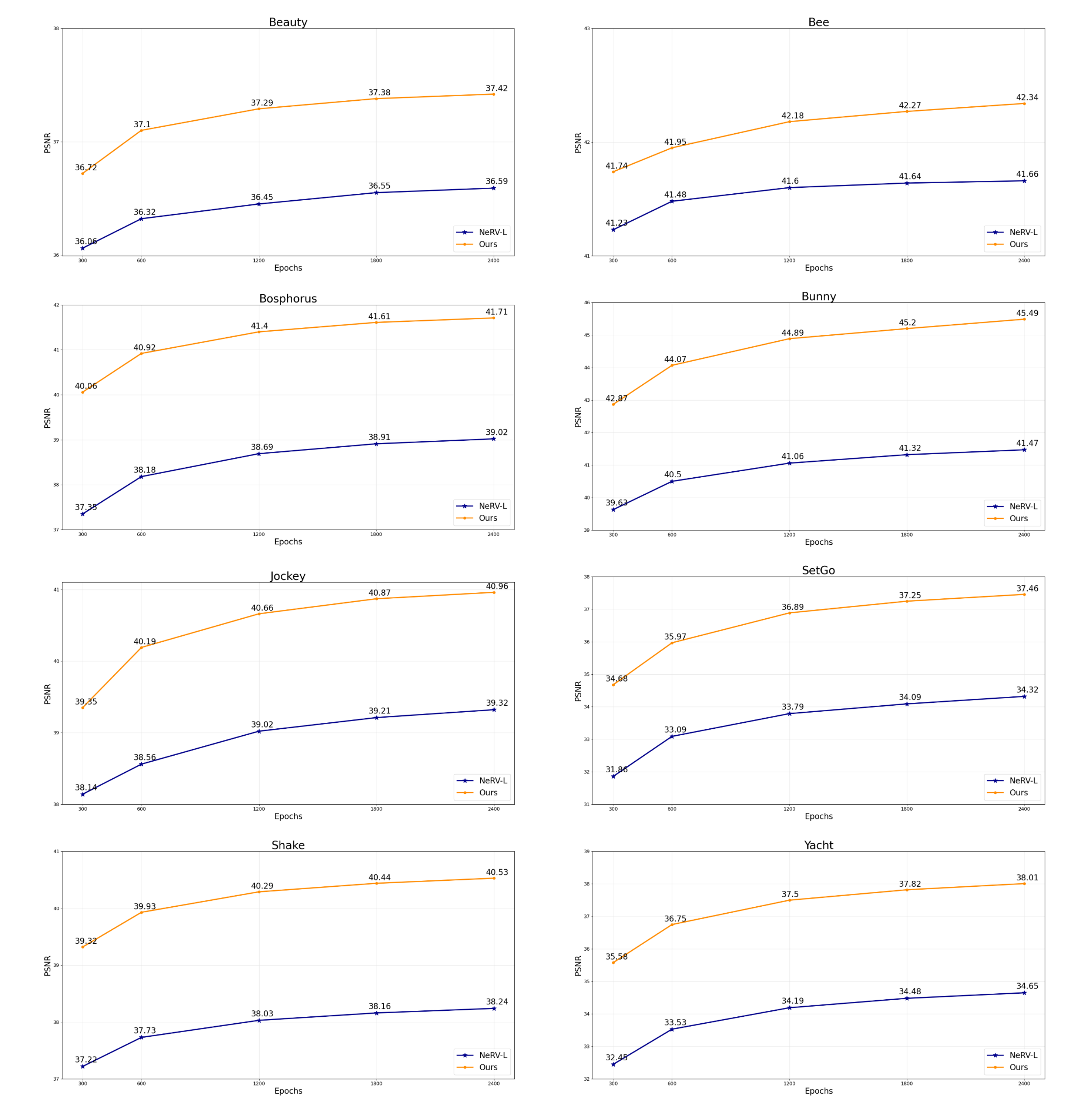}
    \caption{The experiment results of different epochs. Our method's performance at 300 epochs consistently surpass the baseline's results at 2400 epochs on all the video sequences.}
    \label{fig:supp_schedule}
\end{figure}

\setlength{\tabcolsep}{2pt}
\begin{table}[!h]
    \centering
     \caption{Per-video quantitative results of alternative comparison. See Section.~\ref{alternative exp} for a detailed description of each alternative.}
    \resizebox{\textwidth}{!}{
    \begin{tabular}{l|c|cccccccc}
    \toprule
       && \multicolumn{8}{c}{PSNR$\uparrow$}  \\
       {}& Size & Bunny & Beauty & Bosphorus & Bee & Jockey & SetGo & Shake & Yacht \\
       \hline
       NeRV-$C_S$& 5.8M & 35.97 &35.24 & 33.95 &39.88 &34.07 &27.00 &34.98 & 28.67 \\
       Ours-1\dag  & 5.8M &38.94 & 35.81& 36.34 &40.97 &36.95 &30.25 &36.58 &31.44 \\
     NeRV-Split& 7.2M &39.10 &  36.02 &36.29 &40.95 &36.24 &29.64 &36.91 &31.06 \\
     Ours-2\dag & 7.2M & 40.23& 36.27&37.31 &41.31 &37.24 &31.01 &37.65 &32.88 \\
     E-NeRV-MLP & 12M &42.15 &36.66 &39.68 &41.69 &39.23 &34.28 &39.21 &35.43 \\
     E-NeRV-Conv & 12.5M &42.49 & 36.69 & 39.97 &41.70 & 39.28 &34.54 &39.26 &35.50 \\
     \netname & 12.5M & 42.87 & 36.72 & 40.06 & 41.74 & 39.35 & 34.68 & 39.32 & 35.58 \\
    \bottomrule
    \end{tabular}
    }
    \resizebox{\textwidth}{!}{
    \begin{tabular}{l|c|cccccccc}
    \toprule
       && \multicolumn{8}{c}{MS-SSIM$\uparrow$}  \\
       {}& Size & Bunny & Beauty & Bosphorus & Bee & Jockey & SetGo & Shake & Yacht \\
       \hline
       NeRV-$C_S$& 5.8M &0.9843 &0.9446&0.9567&0.9924&0.9509&0.9324&0.9667&0.9212 \\
       Ours-1\dag  & 5.8M &0.9920&0.9508&0.9760&0.9937&0.9710&0.9689&0.9790&0.9590 \\
     NeRV-Split& 7.2M &0.9930 &0.9512 &0.9745 &0.9937&0.9686&0.9629&0.9807&0.9548 \\
     Ours-2\dag & 7.2M & 0.9946&0.9588&0.9801&0.9941&0.9746&0.9736&0.9845&0.9652 \\ 
     E-NeRV-MLP & 12M &0.9964 &0.9633 &0.9898 &0.9945&0.9832& 0.9877&0.9897 &0.9841 \\
     E-NeRV-Conv & 12.5M &0.9966&0.9638&0.9905&0.9946&0.9834&0.9885&0.9899& 0.9843\\
     \netname & 12.5M &0.9969 &0.9887&0.9641&0.9906&0.9946&0.9835&0.9900&0.9846 \\
    \bottomrule
    \end{tabular}
    }
   
    \label{tab:supp_alternative}
\end{table}

\section{More Related Works}

\subsection{Differences between video INRs and video GANs}

Video INR~\cite{sitzmann2020implicit,mehta2021modulated} is a certain type of representation that using network functions to implicitly fit the video sequence. The efficient frame-wise video INRs like NeRV~\cite{chen2021nerv} and E-NeRV proposed in this paper, generate 2-D frame from 1-D latent code, thus having similar architecture compared with video GAN methods.  The difference is that GAN frameworks try to generate diverse videos from different random latent codes, while video INRs try to fit to one individual video precisely. DIfferent latent codes in our case stand for different frame index inputs. On the other hand, video GAN, like recently proposed DiGAN~\cite{yu2021generating} can also adopt video INR as the representation, which inherits the advantage of continuous implicit functions.

\subsection{Discussion of video representation learning}

Another line of related works lies in the self-supervised representation learning. Self-supervised learning~(SSL) methods~\cite{wu2018unsupervised,he2020momentum,chen2020simple,chen2021exploring,wang2021solving,he2022masked,chen2022context} focus on pretext tasks for visual pre-training from unlabelled data, well-studied tasks including contrastive learning~\cite{he2020momentum,chen2020simple,chen2021exploring} and masked image modeling~\cite{he2022masked,chen2022context}. For the video SSL, many works~\cite{wang2015unsupervised,lai2021video,liu2022tcgl,tong2022videomae} also develop specific objectives for video pre-training which aims for the video-related down-stream tasks. For now, the SSL methods utilize network to extract the unique representation of each video, while the video INR tries to fit each video and make the network weight as the representation. Despite of the difference, both methods focus on the representation of video sequence, and how can implicit representation be involved in representation learning, like SSL, is an exciting new direction.

\section{Baseline NeRV Details}

As mentioned in Section.~\ref{datasets and implementation}, for the network structures which are orthogonal to our motivation in NeRV, we follow their original settings. To be more specific, for the MLP, we place a GELU activation function following each linear layer, which in ablation study of NeRV\cite{chen2021nerv} is proved to be superior of other activations. For convolution stages, each NeRV Block also ends with a GELU activation function. Additionally, we don't adopt any of the normalization modules in our model except for the temporal instance normalization. Neither of the Layernorm in transformer and Batchnorm in convolution is adopted. As analyzed in \cite{chen2021nerv}, the normalization layer reduces the over-fitting capability of the neural network, which is contradictory to the training objective of video INR.

\section{More Experiment Results}

\subsection{Alternative Comparison}

Since the experiment results provided in Section.~\ref{alternative exp} are averaged over 8 different video sequences, here we provide the detailed results of each video in Table.~\ref{tab:supp_alternative}. The results show the consistent relative relation among the alternatives on all videos.

\subsection{Ablation Study}

As experiment results provided in Section.~\ref{ablation} are also averaged over 8 different video sequences, the detailed results are provided in Table.~\ref{tab:supp_ablation} for a better look.

\subsection{Schedule Comparison}

In this section we show the comparison of our method and NeRV on all video sequences for different training epochs. As illustrated in Fig.~\ref{fig:supp_schedule}, the PSNR of E-NeRV's output frame images surpasses NeRV's with a much longer $8\times$ schedule. And we can further boost the performance with more epochs. Our method's representation ability shows strong advantage on both the video sequences with dynamic content like ``Bunny'' and the video sequences with more still frames like ``Bee'', which proves the significant improvement of our proposed video INR.

\subsection{Temporal frequency study}

According to the experiments in Section.~\ref{temporal_freq_exp}, we set two different frequency values $b=1.05$ and $b=1.25$, and conduct frame interpolation experiments on all the video sequences. From the results in Table.~\ref{tab:interpolate} we can see the conclusion that adjusting temporal frequency boost both performances on seen and unseen splits still remains, while same adjustment leads to performance drop on seen split for NeRV.

\setlength{\tabcolsep}{2pt}
\begin{table}[t]
    \centering
    \caption{Per-video results of ablation study. As adding the proposed modules, our method's performance increases gradually.}
    \resizebox{\textwidth}{!}{
     \begin{tabular}{l|ccc|cccccccc}
            \toprule
            &&&&\multicolumn{8}{c}{PSNR$\uparrow$} \\
             {} & $\Phi$ & $F_\theta$ & IN & Bunny & Beauty & Bosphorus & Bee & Jockey & SetGo & Shake & Yacht \\
             \midrule
             NeRV-L & - & - & - & 39.63 & 31.86 & 36.06 & 37.35 & 41.23 & 38.14 & 37.22 & 32.45 \\
             \midrule
             Variant 1 & \XSolidBrush & \XSolidBrush & \XSolidBrush & 41.95&36.30&38.97&41.38&38.40&32.81&38.99&34.30 \\
             \midrule
             Variant 2 & \Checkmark & \XSolidBrush & \XSolidBrush & 42.16&36.32&39.29&41.55&38.71&33.16&39.06&34.44 \\
             \midrule
             Variant 3 & \Checkmark & \Checkmark & \XSolidBrush & 42.34&36.52&39.95&41.64&39.02&33.51&39.18&35.40 \\
             \midrule
             \netname & \Checkmark & \Checkmark & \Checkmark & 42.87 & 36.72 & 40.06 & 41.74 & 39.35 & 34.68 & 39.32 & 35.58 \\
             \bottomrule
        \end{tabular}
    }
    
    \resizebox{\textwidth}{!}{
     \begin{tabular}{l|ccc|cccccccc}
            \toprule
            &&&&\multicolumn{8}{c}{MS-SSIM$\uparrow$} \\
             {} & $\Phi$ & $F_\theta$ & IN & Bunny & Beauty & Bosphorus & Bee & Jockey & SetGo & Shake & Yacht \\
             \midrule
             NeRV-L & - & - & - & 0.9931&0.9787&0.9562&0.9825&0.9941&0.9783&0.9883&0.9704 \\
             \midrule
             Variant 1 & \XSolidBrush & \XSolidBrush & \XSolidBrush & 0.9964&0.9570&0.9882&0.9941&0.9806&0.9837&0.9893&0.9805 \\
             \midrule
             Variant 2 & \Checkmark & \XSolidBrush & \XSolidBrush & 0.9966&0.9571&0.9892&0.9944&0.9814&0.9848&0.9896&0.9810 \\
             \midrule
             Variant 3 & \Checkmark & \Checkmark & \XSolidBrush & 0.9967&0.9605&0.9903&0.9945&0.9825&0.9861&0.9898&0.9832 \\
             \midrule
             \netname & \Checkmark & \Checkmark & \Checkmark & 0.9969 &0.9887&0.9641&0.9906&0.9946&0.9835&0.9900&0.9846 \\
             \bottomrule
        \end{tabular}
    }
    \label{tab:supp_ablation}
\end{table}

\begin{table*}[htbp]
    \centering
    \caption{An study of different encoding frequency for temporal interpolation.}
    \setlength{\tabcolsep}{1mm}{
    \begin{tabular}{lcccccccccc}
    \toprule
     Method & split & $b$ & Bunny & Beauty & Bosphorus & Bee & Jockey & SetGo & Shake & Yacht  \\
     \midrule
     \multirow{4}{*}{NeRV} & \multirow{2}{*}{seen} & 1.25 & 39.30 & 36.16 & 37.70 & 41.13 & 38.24 & 32.21 & 37.36 & 32.78 \\
     
     {} & {} & 1.05 & 39.06 & 36.05 & 37.35 & 40.94 & 37.84 & 31.50 & 37.09 & 32.44 \\
     \cmidrule{2-11}
     {} & \multirow{2}{*}{unseen} & 1.25 & 28.58 & 23.98 & 25.65 & 37.08 & 17.27 & 14.69 & 28.05 & 19.83 \\
     
     {} & {} & 1.05 & 33.70 & 26.55 & 29.55 & 39.90 & 18.24 & 15.73 & 30.40 & 22.23 \\
     \midrule
     \multirow{4}{*}{\netname} & \multirow{2}{*}{seen} & 1.25 & 42.52 & 36.96 & 40.71 & 41.72 & 39.78 & 35.44 & 39.58 & 36.32 \\
     
     {} & {} & 1.05 & 42.63 & 37.04 & 40.74 & 41.74 & 40.06 & 35.90 & 39.61 & 36.57 \\
     \cmidrule{2-11}
     {} & \multirow{2}{*}{unseen} & 1.25 & 29.32 & 24.26 & 26.30 & 37.92 & 17.37 & 14.95 & 28.56 & 20.26 \\
     
     {} & {} & 1.05 & 33.77 & 26.41 & 29.10 & 39.67 & 18.24 & 15.91 & 30.57 & 22.31 \\
     \bottomrule
\end{tabular}}
    
    \label{tab:interpolate}
\end{table*}

\section{Visualizations}
\subsection{Reconstructed Frames}

\begin{figure}[!htbp]
    \centering
    \includegraphics[width=\linewidth]{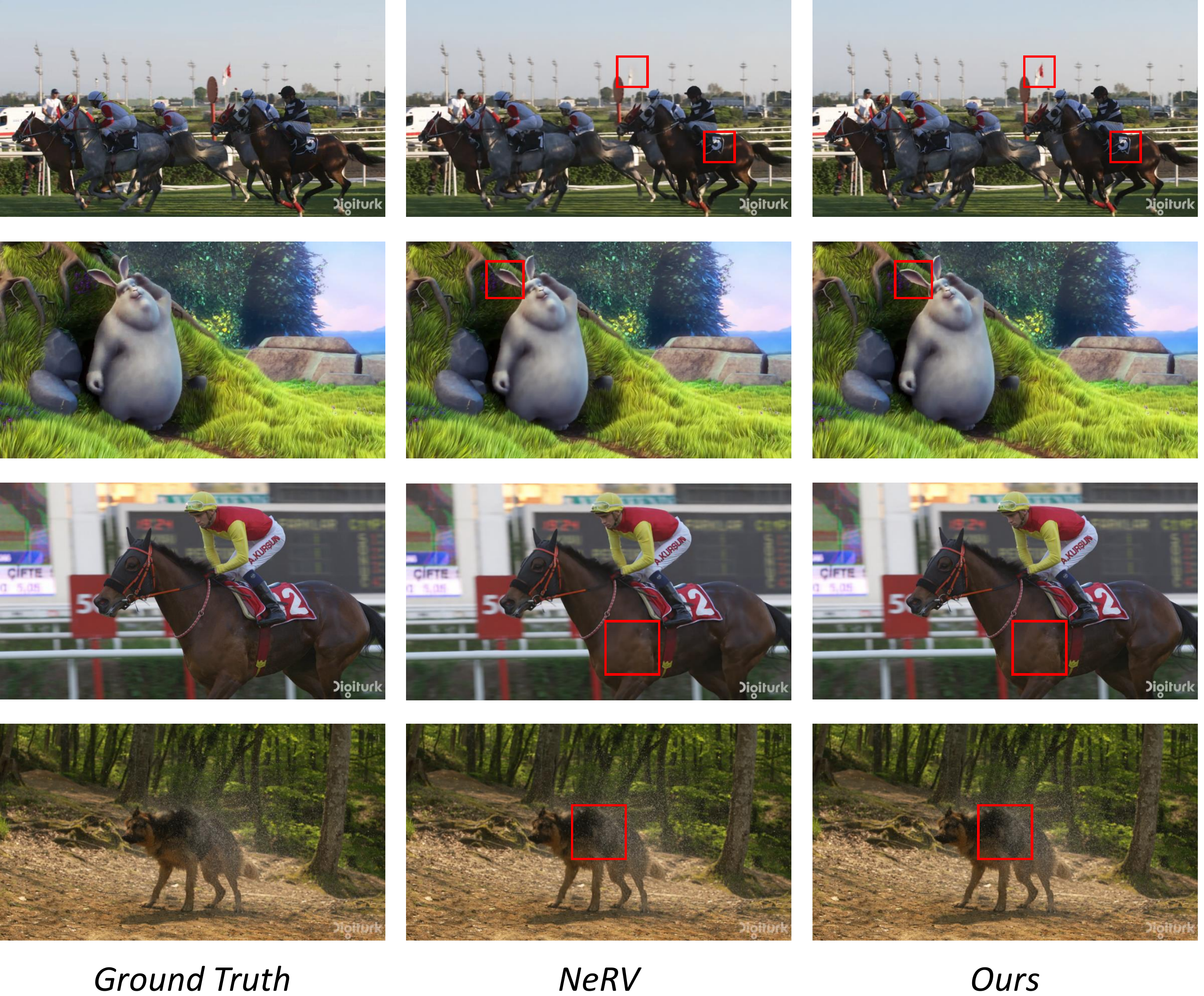}
    \caption{Visualization of reconstructed frames. We use red rectangles to highlight the detailed regions that NeRV fails to synthesis while our method succeeds.}
    \label{fig:supp_vis}
\end{figure}

In this section we show a qualitative comparison between our method and the baseline NeRV-L. As the visualization provided in Fig.~\ref{fig:supp_vis}, despite that NeRV can reconstruct the whole frame image to a good degree, our E-NeRV can further fix some detailed regions. For example, the red flag and elaborate wrinkle in the images of first row.

\subsection{Temporal Interpolation}

\begin{figure}[!htbp]
    \centering
    \includegraphics[width=\linewidth]{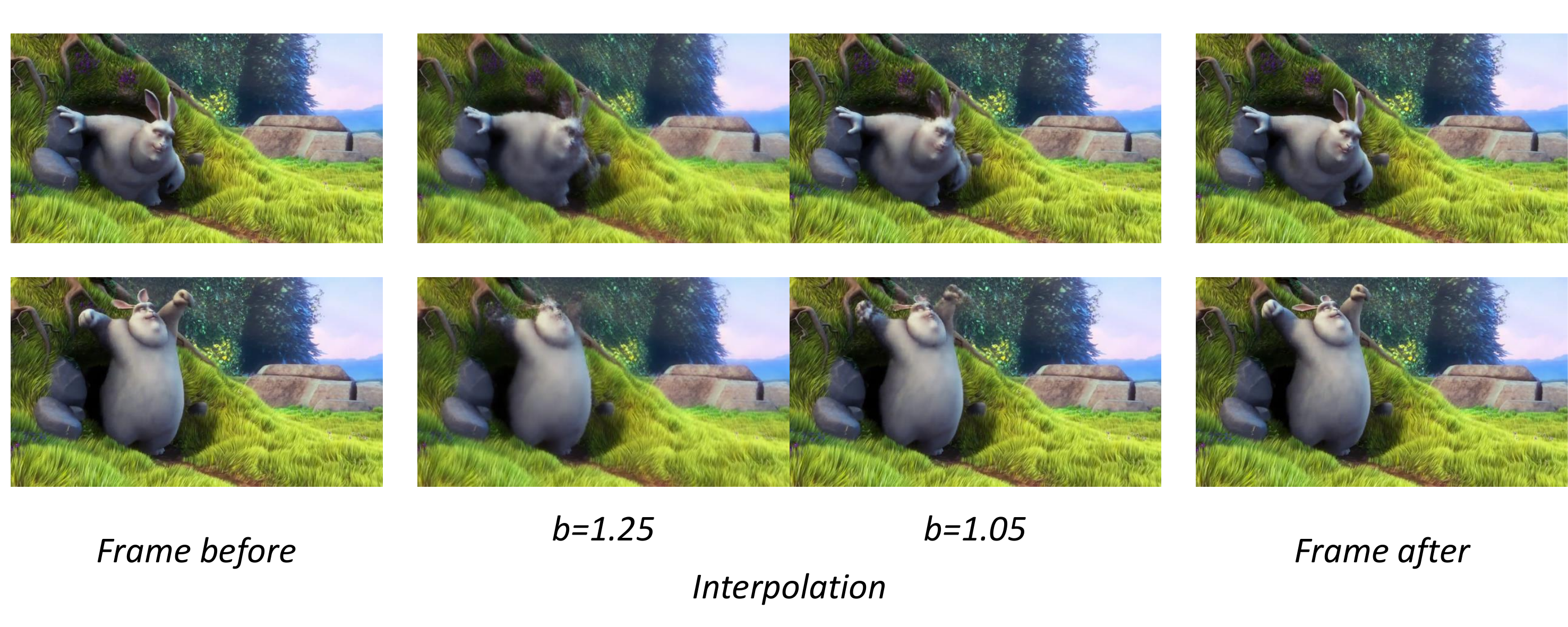}
    \caption{Visualization of frame interpolation. Adjusting the frequency for temporal instance normalization modules provides a more precise interpolation.}
    \label{fig:supp_interp_vis}
\end{figure}

As visualized in Fig.~\ref{fig:supp_interp_vis}, lowering the frequency value for temporal instance normalization can greatly boost E-NeRV's performance on temporal interpolation. The results become more elaborate and less blurry compared to interpolation with original parameters.

